\documentclass[10pt]{article} 
\usepackage[accepted]{tmlr}

\usepackage{xcolor}
\definecolor{mygray}{gray}{.94}
\definecolor{mygreen}{rgb}{0, 0.69, 0.31}
\definecolor{citecolor}{HTML}{0071BC}
\definecolor{linkcolor}{HTML}{ED1C24}
\definecolor{darkergreen}{RGB}{21, 152, 56}
\definecolor{red2}{RGB}{252, 54, 65}
\definecolor{revisecolor}{RGB}{0,0,222}

\usepackage[utf8]{inputenc} 
\usepackage[T1]{fontenc}    
\usepackage{url}            
\usepackage{booktabs}       
\usepackage{amsfonts}       
\usepackage{multirow} 
\usepackage{amsmath}
\usepackage{amssymb}
\usepackage{wrapfig}
\usepackage{graphicx}
\usepackage{multirow}
\usepackage{caption}
\usepackage{pifont}
\usepackage{subcaption}
\usepackage{array}
\usepackage{makecell}
\usepackage{nicefrac}
\usepackage{microtype}
\usepackage{colortbl}
\usepackage{algorithm}
\usepackage{algpseudocode}
\usepackage{listings}

\usepackage{etoolbox}
\makeatletter
\AfterEndEnvironment{algorithm}{\let\@algcomment\relax}
\AtEndEnvironment{algorithm}{\kern2pt\hrule\relax\vskip3pt\@algcomment}
\let\@algcomment\relax
\newcommand\algcomment[1]{\def\@algcomment{\footnotesize#1}}
\renewcommand\fs@ruled{\def\@fs@cfont{\bfseries}\let\@fs@capt\floatc@ruled
  \def\@fs@pre{\hrule height.8pt depth0pt \kern2pt}%
  \def\@fs@post{}%
  \def\@fs@mid{\kern2pt\hrule\kern2pt}%
  \let\@fs@iftopcapt\iftrue}
\makeatother
\usepackage{hyperref}
\newcommand{\ours}{Athena}
\newcommand{\vs}{\textit{vs.}}
\newcommand{\app}{\raise.17ex\hbox{$\scriptstyle\sim$}}
\newcommand{\squishlist}{
\begin{list}{{{\small{$\bullet$}}}}
{\setlength{\itemsep}{3pt}      \setlength{\parsep}{1pt}
\setlength{\topsep}{1pt}       \setlength{\partopsep}{0pt}
\setlength{\leftmargin}{1em} \setlength{\labelwidth}{1em}
\setlength{\labelsep}{0.5em} } }
\newcommand{\squishend}{  \end{list}  }

\newcommand{\yes}{\textcolor{darkergreen}{\ding{52}}}
\newcommand{\no}{\textcolor{red2}{\ding{56}}}

\title{\ours: Enhancing Multimodal Reasoning with Data-efficient Process Reward Models}

\author{\name Shuai Wang\thanks{Work done during the internship at AMD.} \email bit.ybws@gmail.com \\
      \addr Advanced Micro Devices Inc.\\
      The Hong Kong University of Science and Technology (Guangzhou)
      \AND
      \name Zhenhua Liu\thanks{Corresponding authors.} \email zhenhua.liu@amd.com \\
      \addr Advanced Micro Devices Inc.
      \AND
      \name Jiaheng Wei \email jiahengwei@hkust-gz.edu.cn\\
      \addr The Hong Kong University of Science and Technology (Guangzhou)
      \AND
      \name Xuanwu Yin  \email xuanwu.yin@amd.com \\
      \addr Advanced Micro Devices Inc.
      \AND  
      \name Dong Li \email d.li@amd.com \\
      \addr Advanced Micro Devices Inc.
      \AND
       \name Emad Barsoum  \email emad.barsoum@amd.com \\
       \addr Advanced Micro Devices Inc.}

\def\month{05}  
\def\year{2026} 
\def\openreview{\url{https://openreview.net/forum?id=unWmplHccF}} 

\begin{document}
\maketitle
\begin{abstract}
We present \ours-PRM, a multimodal process reward model (PRM) designed to evaluate the reward score for each step in solving complex reasoning problems. Developing high-performance PRMs typically demands significant time and financial investment, primarily due to the necessity for step-level annotations of reasoning steps. Conventional automated labeling methods, such as Monte Carlo estimation, often produce noisy labels and incur substantial computational costs. To efficiently generate high-quality process-labeled data, we propose leveraging prediction consistency between weak and strong completers as a criterion for identifying reliable process labels. Remarkably, \ours-PRM demonstrates outstanding effectiveness across various scenarios and benchmarks with just 5,000 samples. Furthermore, we also develop two effective strategies to improve the performance of PRMs: ORM initialization and up-sampling for negative data. We validate our approach in three specific scenarios: verification for test-time scaling, direct evaluation of reasoning step correctness, and reward ranked fine-tuning. Our \ours-PRM consistently achieves superior performance across multiple benchmarks and scenarios. Notably, when using Qwen2.5-VL-7B as the policy model, \ours-PRM enhances performance by 10.2 points on WeMath and 7.1 points on MathVista for test-time scaling. Furthermore, \ours-PRM sets the state-of-the-art (SoTA) results in VisualProcessBench and outperforms the previous SoTA by 3.9 F1-score, showcasing its robust capability to accurately assess the correctness of the reasoning step. Additionally, utilizing \ours-PRM as the reward model, we develop \ours-7B with reward ranked fine-tuning and outperforms baseline with a significant margin on five benchmarks.
\end{abstract}
\section{Introduction}
\label{sec:intro}
In recent years, Large Language Models (LLMs)~\citep{gpt3,claude3.7,llama2,deepseekv3,qwen2.5} have achieved remarkable success in natural language processing. Building on this, Multimodal Large Language Models (MLLMs)~\citep{llava,li2025llavaonevision,qwen2.5vl,intern2.5vl} have made significant strides in various vision-language tasks, such as visual question answering~\citep{antol2015vqa,mathew2021docvqa} and chart understanding~\citep{chartqa}. Despite their impressive performance, solving complex tasks involving mathematical and multi-step reasoning remains challenging.

To enhance reasoning capabilities, several approaches have been explored, including fine-tuning on long chain-of-thought (CoT) data~\citep{cot,muennighoff2025s1}, offline preference optimization~\citep{internvl_mpo,iterative_reasoning_dpo}, and online reinforcement learning~\citep{shao2024deepseekmath,deepseekr1,hu2025reinforce++,team2025kimi}.  Another promising avenue is test-time scaling (TTS), which involves generating multiple responses from a policy model and selecting the most consistent answer~\citep{wang2023selfconsistency} or the solution with the highest reward using reward models~\citep{cobbe2021training,lightman2024lets,uesato2022solving,snell2025scaling,setlur2025rewarding,wang2024mathshepherd}.

Reward models for TTS mainly include two types: outcome reward models (ORMs)~\citep{cobbe2021training} and process reward models (PRMs)~\citep{lightman2024lets}. ORMs assign the reward score for a given question and its solution, whereas PRMs provide reward scores for each intermediate reasoning step, offering fine-grained feedback. PRMs typically deliver superior performance and better out-of-distribution generalization~\citep{lightman2024lets}. However, obtaining high-quality data with process labels poses significant challenges. PRM800K~\citep{lightman2024lets} involves collecting 800K labeled steps through human annotations, which is time-consuming and requires skilled annotators, particularly for complex multi-step reasoning and mathematical tasks. Math-Shepherd~\citep{wang2024mathshepherd} proposes an automated labeling method using Monte Carlo (MC) estimation. It defines the quality of an intermediate step as its potential to reach the correct answer. MC estimation for step-level labels is typically performed by sampling numerous reasoning trajectories using a language model, referred to as the \textit{completer}, to estimate the probability of arriving at the correct answer. However, this approach involves significant computational overhead. Besides, another shortcoming of MC-based estimation methods is that estimated step-level labels are inevitably noisy.

\begin{figure}[t]
    \centering  
    \includegraphics[width=0.99\linewidth]{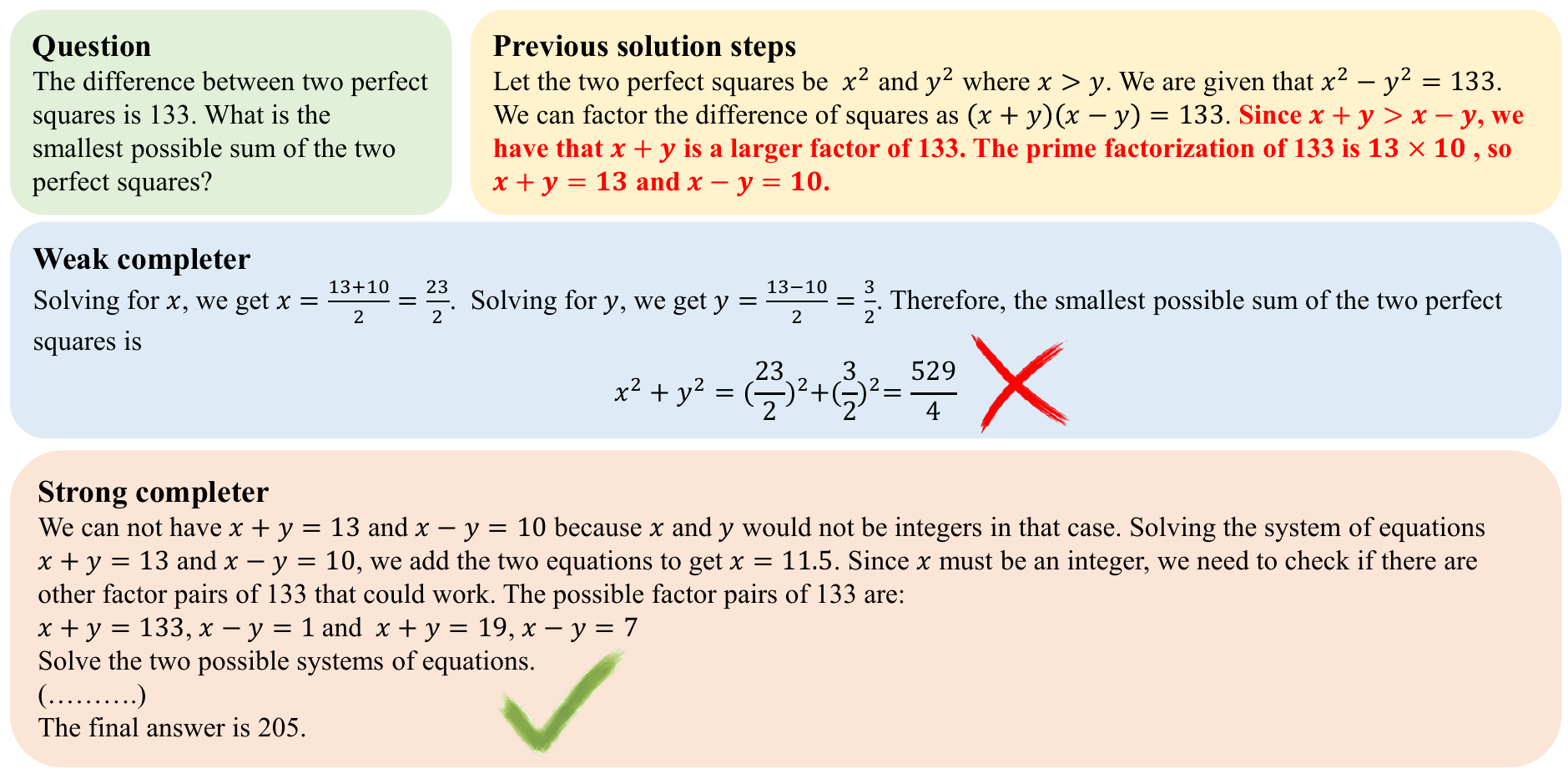}    
    \caption{Illustration of different completers under the same question and solution steps. Even if given \textcolor{red}{wrong intermediate steps}, the strong completer still reaches the final answer while the weak completer fails. We omitted some intermediate steps in the figure for simplicity.}
    \label{fig:illustration_completer}
\end{figure}

We aim to address the above challenges: reducing computational costs and mitigating noisy step labels. We first find that the accuracy of step labels estimated by MC-based methods is influenced by the reasoning capability of the completer. A strong completer can arrive at the correct answer despite incorrect intermediate steps as shown in Figure~\ref{fig:illustration_completer}, whereas a weak completer may struggle even with correct intermediate steps. The estimation of MC-based methods is biased toward the completers we use. However, the correctness of the intermediate step should not depend on the completer. Based on this insight, \textbf{we propose using both weak and strong completers to estimate step labels, retaining only those steps where labels generated by both completers are consistent to remove the bias caused by completers}. This approach improves the quality of the step label. Empirical results demonstrate that a small set of high-quality labels ($\sim$5K) achieves impressive performance compared to large-scale data ($\sim$ 300K) labeled by vanilla MC-based estimation~\citep{wang2024mathshepherd}. In addition, our approach requires only \textbf{1/45} of the GPU hours for data synthesis and \textbf{1/60} of the GPU hours for reward model training, significantly lowering computational costs.

After acquiring high-quality datasets, we explore two effective strategies for training PRMs: \textbf{initialization from ORMs} and \textbf{up-sampling data with negative labels}. PRMs are typically fine-tuned from pre-trained foundation models, such as LLaMA-3.1-8B-Instruct~\citep{llama3} in RLHF-workflow~\citep{rlhf_workflow}. Previous studies indicate that ORMs possess some ability to assess intermediate step correctness through confidence variation~\citep{lu2024autopsv} or parameterization~\citep{cui2025process,yuan2024implicitprm}. Inspired by this, we find that initializing PRMs from ORMs significantly boosts performance, as ORMs trained on large-scale \textit{sample-level} data serve as pre-training with coarse-grained data. And PRMs can be considered fine-tuning on high-quality, fine-grained \textit{step-level} data. Additionally, we show that \textit{label imbalance} exists in process-labeled solutions and we propose to up-sample data with negative step labels to address this.

Building on these methodologies, we develop our outcome reward model \ours-ORM and process reward model \ours-PRM. Leveraging \ours-PRM, we introduce \ours-7B with reward ranked fine-tuning. We validate our approach across three scenarios: 1) test-time scaling ~\citep{snell2025scaling}: where \ours-PRM ranks multiple outputs generated by policies under the Best-of-N evaluation; 2) direct evaluation of the correctness of reasoning steps using \ours-PRM; 3) reward ranked fine-tuning~\citep{dong2023raft}: where \ours-PRM ranks outputs sampled from the current policy, utilizing the highest reward response for policy model fine-tuning. In the TTS scenario, we evaluate \ours-PRM on seven multimodal math and reasoning benchmarks with three different policy models ranging from 7B to 72B, demonstrating significant improvements in reasoning abilities. For instance, on the WeMath benchmark~\citep{wemath}, \ours-PRM enhances the zero-shot baseline by \textbf{10.2} points using Qwen2.5-VL-7B~\citep{qwen2.5vl} as the policy model. Furthermore, \ours-PRM excels in text-only benchmarks, achieving an \textbf{8.9} points improvement on a challenging math benchmark~\citep{hendrycks_math} with Mistral-8B. To assess its ability to judge intermediate reasoning step correctness, \ours-PRM is evaluated on the VisualProcessBench~\citep{wang2025visualprm}, showcasing strong performance and outperforming VisualPRM-8B~\citep{wang2025visualprm}, an open-source multimodal PRM, and proprietary models as judge. In the reward ranked fine-tuning scenario, our fine-tuned model \ours-7B, based on Qwen2.5-VL-7B~\citep{qwen2.5vl}, significantly enhances the policy model's reasoning capabilities across seven math and reasoning benchmarks.

Our contributions are summarized as follows:
\squishlist
\item We propose using prediction consistency between weak and strong completers to filter noisy process labels, enhancing the quality of the automated-labeled process data. The high-quality process labeled data shows surprising data efficiency for training PRMs.
\item We introduce two effective strategies to improve the performance of PRMs: ORM initialization and negative data up-sampling.
\item We develop our reward models \ours-ORM and \ours-PRM, and evaluate them under the Best-of-N setting on nine math and reasoning benchmarks across various model sizes and families, demonstrating the effectiveness of our approach. Moreover, \ours-PRM achieves state-of-the-art (SoTA) performance in assessing the correctness of intermediate steps directly on the VisualProcessBench~\citep{wang2025visualprm}.
\item Leveraging \ours-PRM, we introduce \ours-7B, a MLLM with exceptional capabilities fine-tuned using reward ranked fine-tuning. Extensive experiments highlight the superiority of \ours-7B across diverse benchmarks.
\squishend

\section{Related Work}
\textbf{Test-time scaling.}~ While scaling data and model parameters during training has been extensively studied~\citep{kaplan2020scaling,hoffmann2022training}, scaling computation at test-time has garnered interest more recently~\citep{snell2025scaling,setlur2025rewarding}. TTS enables language models to leverage additional computational resources when faced with challenging questions. Our study primarily focuses on parallel scaling~\citep{tts-survey}, wherein language models generate multiple outputs simultaneously and aggregate them into a final answer. This aggregation can involve selecting the most common outcome~\citep{wang2023selfconsistency} or using reward models~\citep{snell2025scaling} to rank solutions, selecting the one with the highest reward, as demonstrated in our Best-of-N settings.

\textbf{Process reward models.}~ Reward models are crucial in reinforcement learning~\citep{deepseekr1,shao2024deepseekmath,wang2024mathshepherd,instruct_gpt,bai2022training,rlhf_workflow} and test-time scaling~\citep{setlur2025rewarding,lightman2024lets,wang2024mathshepherd,cobbe2021training,snell2025scaling}. Outcome reward models~\citep{cobbe2021training} assign a scalar value to question-response pairs, whereas process reward models evaluate each step, typically achieving superior performance and generalization~\citep{lightman2024lets,OmegaPRM,setlur2025rewarding,li2025process_q_ranking,multi-rm}. While better performance and generalization are achieved, collecting data to train PRMs is challenging. PRM800K~\citep{lightman2024lets} is the first open-source process supervision dataset annotated by humans, but scaling it is challenging due to the time and skill required for annotating reasoning steps. Math-Shepherd~\citep{wang2024mathshepherd} proposes an automated method for estimating intermediate step correctness using Monte Carlo estimation, though it generates some incorrect labels and demands extensive computational resources. Our work explores strategies to reduce computational costs while improving process label accuracy.

\section{Method}
\label{sec:method}
In this section, we first introduce basic concepts, including ORMs and PRMs in Sec.~\ref{sec:method_reward}. Next, we present our method for constructing a high-quality PRM training set, along with practical training strategies to improve PRM performance in Sec.~\ref{subsec:auto_prm} and \ref{sec:method_strategy}. The data curation process for training is detailed in Sec.~\ref{sec:method_data}. Finally, we discuss three application scenarios of PRMs in Sec.~\ref{sec:method_application}.
\subsection{Preliminaries:Reward Models for Mathematical Problem}
\label{sec:method_reward}
\textbf{ORMs.}~ Given a mathematical problem $x$ with golden answer $y^{*}$ and its solution $a$ generated by policy $\pi$, i.e., $a\sim \pi \left(\cdot \mid x \right)$, ORMs assign a reward $r\left(x,a\right)\in \left(0,1\right)$ to reflect the correctness of the solution $a$. To train ORMs, we take inputs as $(x,a)$ and predict correctness $\delta(y,y^{*})$ of prediction with following loss function:
\begin{equation}
    \mathcal{L}_{ORM}=-\left(\delta\left(y,y^{*}\right)\cdot \log r\left(x,a\right) + \left(1-\delta\left(y,y^{*}\right) \right)\cdot\log \left(1-r\left(x,a\right)\right)\right),
    \label{eq:orm}
\end{equation}
where $\delta\left(y,y^{*}\right)=1$ if $y=y^{*}$, otherwise $\delta\left(y,y^{*}\right)=0$.

\textbf{PRMs.}~ Different from ORMs, PRMs aim to predict the correctness of \textit{each step} in the solution. Specifically, we sample response $a\sim \pi \left(\cdot \mid x \right)$. A response $a$ usually consists of multiple reasoning steps separated by a delimiter (e.g. \textbackslash n\textbackslash n), a.k.a $a=(a_1,a_2,\dots,a_K)$ when $K$ is the number of reasoning steps. To predict the correctness of each step, we train PRMs with the cross-entropy loss for each step in the solution. Formally, we get the following loss function:
\begin{equation}
    \mathcal{L}_{PRM}=-\left(\sum_{i=1}^K \delta_i\cdot\log r\left(x,a_i\right) + \left(1-\delta_i \right)\cdot\log \left(1-r\left(x,a_i\right)\right)\right),
    \label{eq:prm}
\end{equation}
where $\delta_i=1$ if the step $a_i$ is correct, otherwise $\delta_i=0$.

\subsection{Automated Process Label Annotation}
\label{subsec:auto_prm}
Compared with ORMs, PRMs usually achieve better performance and exhibit stronger out-of-distribution generalization~\citep{lightman2024lets}. However, collecting high-quality data with process labels is challenging. Human annotation provides accurate supervision signals~\citep{lightman2024lets}, but it requires expensive human labeling and difficult to scale up. Math-Shepherd~\citep{wang2024mathshepherd} proposes to use a Monte Carlo (MC) sampling method to estimate the correctness of each step without human supervision. Specifically, to estimate the correctness of step $a_i$, we use a \textit{completer} $\phi$ to finalize the reasoning process from step $a_i$, and get the final answer $y$. We repeat sampling $T$ times and get corresponding answers $\mathcal{Y}=\{y_j\}_{j=1}^T$. There are two ways to estimate the correctness of $a_i$: soft estimation and hard estimation.

For soft estimation, we assume that the frequency of getting the correct answer could dictate the quality of a step:
\begin{equation}
    \delta_{i}^{soft} = \frac{\sum_{j=1}^T \mathbb{I}(y_j=y^{*})}{T}.
\end{equation}
For hard estimation, we assume that the correct answer is right when the step could reach the answer in $T$ samples:
\begin{equation}
    \delta_{i}^{hard} = \begin{cases}
        1 & \text{if } \exists y_j \in \mathcal{Y}, y_j = y^*, \\
        0 & \text{otherwise.}
    \end{cases}
\end{equation}
In this paper, we mainly discuss the hard estimation for PRMs because it allows us to use standard language modeling loss to train PRMs and eliminates the adjustments of the training pipeline.

MC-based estimation methods provide an automated and scalable labeling strategy for intermediate reasoning steps. However, automatically labeled data often contain incorrect labels, and MC-based methods typically incur substantial computational costs due to extensive sampling requirements. For instance, when processing a solution $a$ with $K$ steps, the \textit{completer} $\phi$ must generate $T\times K$ solutions, which produces $T\times K$ times computation cost compared with ORMs.

To enhance the accuracy of the labeling process and reduce the computational burden of data synthesis, our central approach is to utilize a smaller set of high-quality data with process labels. We improve the quality of automatically labeled process labels by introducing consistency filtering between weak and strong completers. The rationale is straightforward: the results of hard estimation depend on the number of samples $T$, the difficulty of the problem $x$, and the ability of the completer $\phi$. A strong completer can arrive at the correct answer even when provided with incorrect steps, as shown in Figure~\ref{fig:illustration_completer}, whereas a weak completer struggles to find the final answer even when given correct steps. We aim to enhance the quality of process labels through prediction consistency between weak and strong completers. Specifically, we only use steps whose assigned labels are consistent between different completers.

To validate our method, we conducted a simple experiment using the PRM800K dataset~\citep{lightman2024lets}. We employed Mistral-7B-Instruct~\citep{jiang2023mistral7b} as the weak completer $\phi_{w}$ and Qwen2.5-72B-Instruct~\citep{qwen2.5} as the strong completer $\phi_{s}$, setting the number of samples $T=8$ for MC-based estimation. We randomly sampled 50 queries and reported the accuracy of estimated process labels:
\begin{center}
    \begin{tabular}{cccc}
        & Weak Completer $\phi_{w}$ & Strong Completer $\phi_{s}$ & Consistency Filter \\ 
        \midrule
     Accuracy (\%) &   78.2 & 83.4 & \textbf{94.1} \\
    \end{tabular}
\end{center}

Our experiments show that \textbf{label quality improves significantly after applying the consistency filter}. Empirically, we demonstrate that \textbf{a small number of high-quality labels is sufficient to achieve superior performance}, thereby reducing computational costs compared to the vanilla MC-based estimation~\citep{wang2024mathshepherd}. Results of more strategies and combination of completers are provided in Table~\ref{tab:exp_completer} of Appendix~\ref{sec:appendix_choice_completer}.

\subsection{Additional Strategies for Training PRMs}
\label{sec:method_strategy}
We propose two strategies for enhancing the performance of PRMs: initialization from ORMs and negative data up-sampling. First, initializing PRMs with ORMs trained on large-scale sample-level annotated data improves performance. Previous work indicates that ORMs possess a certain capacity to assess the correctness of intermediate steps~\citep{cui2025process,lu2024autopsv,yuan2024implicitprm}. We empirically show that PRMs benefit from a few high-quality examples when initialized with ORMs. Large-scale sample-level annotation data provides weaker supervision, with outcome supervision acting as a simplified form of process supervision. Training ORMs on large-scale sample-level annotations serves as a ``pre-training'', while training PRMs from pre-trained ORMs acts as``fine-tuning'' with high-quality process-level annotated data. 

Secondly, our findings reveal that \textit{label imbalance} is prevalent in most process-labeled datasets, such as PRM800K~\citep{lightman2024lets} and our synthetic data, where correct steps are more common than errors. We provide a detailed distribution of process labels in Table~\ref{tab:label_stats}. We up-sample data with negative labels to tackle this problem and empirical results show that up-sampling data with negative labels improves performance with minimal additional computation.

\begin{figure}[t]
    \centering      
\includegraphics[width=0.99\linewidth]{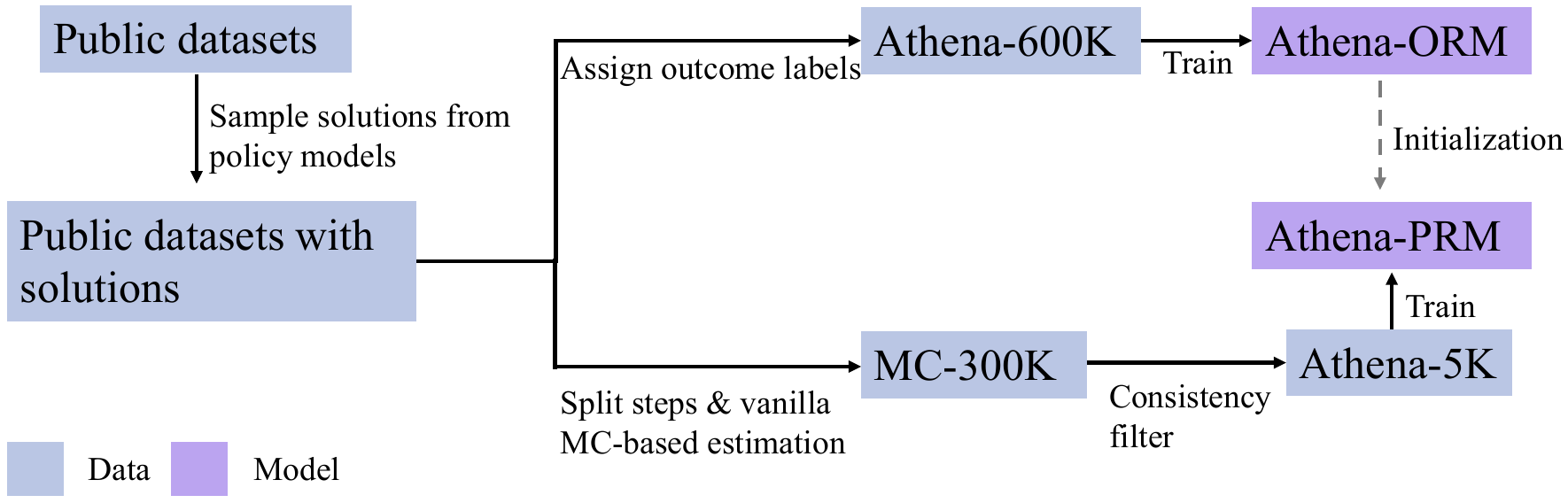}    
    \caption{The pipeline of our datasets and models.}
    \label{fig:data}
\end{figure}

\subsection{Training Data Curation}
\label{sec:method_data}
In this section, we detail the curation process of process labeled datasets and the pipeline is shown in Figure~\ref{fig:data}.

To construct a diverse and high-quality dataset, we first gather data from various public multimodal and text-only datasets. For multimodal datasets, we collect from MathV360k~\citep{shi2024math}, UniGeo~\citep{chen2022unigeo}, Geometry3k~\citep{geo3k}, CLEVR-Math~\citep{lindstrom2022clevr}, ScienceQA~\citep{scienceqa}, GeomVerse~\citep{geomverse}, GeoQA-plus~\citep{cao2022augmented}, and DocVQA~\citep{mathew2021docvqa}. For text-only datasets, we utilize GSM8K~\citep{cobbe2021training}, MATH~\citep{hendrycks_math}, and NuminaMath-1.5~\citep{numina_math_datasets}.

If the original dataset includes different splits, we use only the training portion to prevent test set contamination. We remove all judgment/proof problems and convert multiple-choice questions to open-ended ones to deter policy models from ``guessing'' answers. Then, for each query, we sample 8 responses from the policy model. We prompt all models to generate solutions whose steps are separated by \texttt{\textbackslash n\textbackslash n} and the final answer is included in \texttt{\textbackslash boxed\{\}}. We use \texttt{math--verify}\footnote{\url{https://github.com/huggingface/Math-Verify}} package to compare extracted answer and groundtruth. To ensure dataset diversity, we employ multiple models as policies, including InternVL2.5-8B/78B~\citep{intern2.5vl}, Qwen2.5-VL-7/72B-Instruct~\citep{qwen2.5vl}, LLaVA-OneVision-7B/72B~\citep{li2025llavaonevision} for multimodal datasets, and Qwen2.5-7/72B-Instruct~\citep{qwen2.5} for text-only datasets.

Next, we apply carefully designed filtering rules to all responses. We exclude responses with too many or too few tokens and eliminate those exhibiting repetitive patterns. We employ $n$-gram deduplication to remove similar responses and enhance diversity. Ultimately, we obtain approximately 600K queries with corresponding responses. We parse answers from responses and assign correctness labels of 0 or 1 to each query-response pair. We refer to this dataset as \ours-600K and employ it to train our outcome reward model \ours-ORM.

For training PRMs, we generate process labels using the proposed method introduced in Sec.~\ref{subsec:auto_prm}. We use Qwen2.5-VL-3B~\citep{qwen2.5vl} as the weak completer $\phi_{w}$ and Qwen2.5-VL-72B~\citep{qwen2.5vl} as the strong completer $\phi_{s}$. Finally, we collect approximately 5K samples for training our process reward model \ours-PRM. We also compare with PRMs trained using approximately 300K samples estimated by the vanilla MC-based method outlined in~\citet{wang2024mathshepherd}. We denote these datasets as \ours-5K and MC-300K, respectively, and compare their performance and computational cost in Sec.~\ref{sec:exp_analysis}.

\subsection{Application Scenarios}
\label{sec:method_application}
Following reward model training, we assess the performance of ORMs and PRMs in three scenarios: test-time scaling~\citep{snell2025scaling}, direct judgment, and reward ranked fine-tuning~\citep{dong2023raft}.

\textbf{Verification for test-time scaling.}~  We adopt a Best-of-N evaluation paradigm. Specifically, given a problem $x$ in the test set, we sample $N$ solutions from the policy $\pi$. All solutions are scored using a reward model, and we choose the solution with the highest score. For ORMs, we directly use the outputs from ORMs as the reward of solutions. For PRMs, the minimum reward across all steps is used to rank all solutions following~\citep{wang2024mathshepherd,lightman2024lets}. We test different choices as the reward for solutions in Appendix~\ref{sec:appendix_reward_choice}.

\textbf{Direct judgment for reasoning steps.}~ Besides verification at test-time under the Best-of-N setting, \ours-PRM can also be used to directly identify erroneous steps in the mathematical reasoning process. Given a solution $a$ with $K$ steps as input, \ours-PRM outputs the correctness score $\{\delta_i\}_{i=1}^K$ for each step $\{a_i\}_{i=1}^K$ in a single forward pass.

\textbf{Response ranking for reward ranked fine-tuning.}~  We explore the use of PRMs for data synthesis in reward ranked fine-tuning~\citep{dong2023raft,yuan2023scaling,scale_selftraining,zeng2025bstar}. Using the current policy $\pi$, we generate $M=8$ solutions for an input problem $x$. Subsequently, we filter out solutions with incorrect answers and apply deduplication to remove highly similar responses, enhancing the diversity of the remaining solutions. We retain queries where 2 to 6 out of 8 responses are correct, excluding those with too few or too many correct answers, as such cases are either too easy or too challenging for the current policy $\pi$. Following this filtering step, we use \ours-PRM to score all solutions for each query and select the solution with the highest reward as the corresponding label. The policy $\pi$ is fine-tuned using the synthetic data generated as described above.

\section{Experiments}
\label{sec:experiment}
To comprehensively evaluate the proposed method, we assess \ours-ORM and \ours-PRM under the Best-of-N setting on seven multimodal benchmarks and two text-only benchmarks in Sec.~\ref{sec:bon_exp}. Furthermore, we evaluate \ours-PRM on VisualProcessBench, comparing its performance with MLLMs as judges and a recent open-source multimodal PRM, VisualPRM-8B~\citep{wang2025visualprm}, as described in Sec.~\ref{sec:exp_vlp}. We assess the capability of \ours-7B on multimodal math and reasoning tasks in Sec.~\ref{sec:exp_finetune}. To validate the effectiveness of our designs for \ours-PRM, an in-depth analysis is provided in Sec.~\ref{sec:exp_analysis}.

\subsection{Best-of-N Evaluation}
\label{sec:bon_exp}
\textbf{Benchmarks.}  We evaluate \ours-ORM and \ours-PRM across multiple multimodal mathematical and logical reasoning benchmarks, including MathVista~\citep{lu2024mathvista}, MathVision~\citep{mathvision}, MathVerse~\citep{zhang2024mathverse}, WeMath~\citep{wemath}, DynaMath~\citep{zou2025dynamath}, LogicVista~\citep{xiao2024logicvista}, and MMMU~\citep{yue2024mmmu}.
To validate the effectiveness of \ours-ORM and \ours-PRM in the text-only scenario, we conduct experiments on the GSM8K~\citep{cobbe2021training} and MATH~\citep{hendrycks_math} under the BoN settings. Further details can be found in Appendix~\ref{sec:appendix_benchmark_detail}.

\textbf{Settings.} We employ \ours-ORM and \ours-PRM as the reward model for BoN evaluation, sampling $N=8$ solutions for every problem by default. For decoding, we set the temperature to 0.8 and use nucleus sampling~\citep{top_p} with top-$p$ set to 0.9, and report the average of five runtimes. For multimodal benchmarks, we choose Qwen2.5-VL-7B/72B~\citep{qwen2.5vl} and InternVL2.5-8B~\citep{intern2.5vl} as policy models.\footnote{All models used in this paper are ``instruct'' version. For simplicity, we ignore ``instruct'' in the text.}  For text-only datasets, we choose Qwen2.5-7/72B~\citep{qwen2.5} and Ministral-8B\footnote{\url{https://huggingface.co/mistralai/Ministral-8B-Instruct-2410}} as policy models. Zero-shot and self-consistency~\citep{wang2023selfconsistency} serve as our baselines, alongside a recent public multimodal process reward model, VisualPRM-8B~\citep{wang2025visualprm}.

\begin{table}[t]
\centering
\caption{\textbf{Results on seven multimodal reasoning benchmarks under Best-of-N (N=8) evaluation}. $\dagger$ denotes that the results are from~\citet{wang2025visualprm}. We \colorbox{mygray}{mark} our results and highlight the \textcolor{red2}{\textbf{best}} result.}
\resizebox{0.99\textwidth}{!}{
\begin{tabular}{l*{7}{c}}
\toprule
 & \textbf{WeMath} & \textbf{MathVista} & \textbf{MathVision} & \textbf{MathVerse} & \textbf{DynaMath} & \textbf{MMMU} & \textbf{LogicVista}  \\
\midrule
         Qwen2.5-VL-7B~\citep{qwen2.5vl} & 36.2   & 68.1 & 25.4 & 41.1  & 21.8 & 58.0 & 47.9   \\
         Self-consistency~\citep{wang2023selfconsistency} & 44.7   & 71.6 & 28.6 & 43.7  & 22.9 & 60.1 & 49.5   \\
        $\text{VisualPRM-8B}^{\dagger}$~\citep{wang2025visualprm} & 39.8 & 70.3 & 31.3 & 44.3 & 23.0 & 58.6 & 48.3 \\
         \rowcolor{mygray} \ours-ORM       & 45.1   & 72.8 & 29.8 & 44.1  & 23.1 & 62.7 & 51.3   \\
           \rowcolor{mygray} \ours-PRM       & \textcolor{red2}{\textbf{46.4}}  &  \textcolor{red2}{\textbf{75.2}} &  \textcolor{red2}{\textbf{32.5}} & \textcolor{red2}{\textbf{46.3}}  & \textcolor{red2}{\textbf{23.4}} &  \textcolor{red}{\textbf{63.8}} &   \textcolor{red2}{\textbf{53.0}} \\
\midrule
    InternVL2.5-8B~\citep{intern2.5vl} & 23.5   & 64.5 & 17.0 & 22.8 & 9.4  & 56.2 & 36.0    \\
     Self-consistency~\citep{wang2023selfconsistency}     & 28.4   & 66.1 & 21.1 & 24.7 & 13.8 & 57.8 & 40.2    \\
       $\text{VisualPRM-8B}^{\dagger}$~\citep{wang2025visualprm} & \textcolor{red2}{\textbf{36.5}} & 68.5 & \textcolor{red2}{\textbf{25.7}} & \textcolor{red2}{\textbf{35.8}}  & 18.0 & 60.2 & 43.8 \\ 
       \rowcolor{mygray} \ours-ORM       & 28.6   & 66.9 & 22.0 & 25.8 & 15.2 & 59.1 & 40.8    \\
       \rowcolor{mygray} \ours-PRM       & 30.1   & \textcolor{red2}{\textbf{71.4}} & 23.4 & 26.1 &  \textcolor{red2}{\textbf{18.7}} & \textcolor{red2}{\textbf{60.3}} &   \textcolor{red2}{\textbf{44.4}} \\ 
\midrule
   Qwen2.5-VL-72B~\citep{qwen2.5vl} & 49.1  & 74.2 & 39.3 & 47.3 & 35.9 & 70.2 & 55.7     \\
   Self-consistency~\citep{wang2023selfconsistency}  & 54.8  & 77.0 & 43.1 & 50.8 & 37.6 & 71.1 & 59.6  \\
     \rowcolor{mygray}  \ours-ORM      & 55.6  & 77.8 & 43.0 & 51.2 & 39.6 & 72.3 & 60.1 \\
     \rowcolor{mygray}  \ours-PRM      &   \textcolor{red2}{\textbf{58.7}} &  \textcolor{red2}{\textbf{79.1}} &  \textcolor{red2}{\textbf{44.8}} &  \textcolor{red2}{\textbf{54.6}}  &  \textcolor{red2}{\textbf{42.5}} &  \textcolor{red2}{\textbf{75.8}} &  \textcolor{red2}{\textbf{60.9}} \\    
\bottomrule
\end{tabular}
}
\label{tab:exp_bon}
\end{table}
\begin{table}[t]
\centering
\caption{\textbf{Results on text-only math benchmarks under Best-of-N (N=8) evaluation}. $\dagger$ denotes that the results are from~\citet{wang2025visualprm}. We \colorbox{mygray}{mark} our results and highlight the \textcolor{red2}{\textbf{best}} result.}

\begin{minipage}{0.3\textwidth} 
\centering
\caption*{(a) Qwen2.5-7B}
\resizebox{\textwidth}{!}{
\begin{tabular}{lcc}
\toprule
                                       & \textbf{GSM8K} & \textbf{MATH} \\
\midrule
Qwen2.5-7B~\citep{qwen2.5}             & 91.6  & 75.5  \\
Self-consistency~\citep{wang2023selfconsistency}    & 93.4  & 79.9  \\
$\text{VisualPRM-8B}^{\dagger}$~\citep{wang2025visualprm} & 94.5 & 81.6 \\
\rowcolor{mygray} \ours-ORM                 &  94.0  & 81.1   \\
\rowcolor{mygray} \ours-PRM                 &   \textcolor{red2}{\textbf{94.8}}  &  \textcolor{red2}{\textbf{82.4}}  \\
\bottomrule
\end{tabular}}
\end{minipage}
\hspace{0.03\textwidth} 
\begin{minipage}{0.3\textwidth}
\centering
\vspace{-2.5mm}
\caption*{(b) Ministral-8B}
\resizebox{\textwidth}{!}{
\begin{tabular}{lcc}
\toprule
                                       & \textbf{GSM8K} & \textbf{MATH} \\
\midrule
Ministral-8B                           & 85.6  & 54.5  \\
Self-consistency~\citep{wang2023selfconsistency}    & 90.7  & 62.0  \\
\rowcolor{mygray} \ours-ORM        &  91.4  &  63.2  \\
\rowcolor{mygray} \ours-PRM       & \textcolor{red2}{\textbf{93.1}}  & \textcolor{red2}{\textbf{65.4}}  \\
\bottomrule
\end{tabular}}
\end{minipage}
\hspace{0.03\textwidth}
\begin{minipage}{0.3\textwidth}
\centering
\caption*{(c) Qwen2.5-72B}
\resizebox{\textwidth}{!}{
\begin{tabular}{lcc}
\toprule
                                       & \textbf{GSM8K} & \textbf{MATH} \\
\midrule
Qwen2.5-72B~\citep{qwen2.5}            & 95.8  & 83.1  \\
Self-consistency~\citep{wang2023selfconsistency}    & 96.0  & 86.0  \\
$\text{VisualPRM-8B}^{\dagger}$~\citep{wang2025visualprm} & 96.5 & 85.2 \\
\rowcolor{mygray} \ours-ORM                 & 96.0  & 86.2   \\
\rowcolor{mygray} \ours-PRM                 & \textcolor{red2}{\textbf{97.0}}  & \textcolor{red2}{\textbf{87.4}}  \\
\bottomrule
\end{tabular}}
\end{minipage}
\label{tab:exp_bon_text}
\end{table}

\textbf{Training details.}  To train ORMs and PRMs, we fine-tune Qwen2.5-VL-7B on our dataset as introduced in Sec.~\ref{sec:method_data}. We use the AdamW optimizer~\citep{adamw} with the weight decay as 0 and set the learning rate as $1e^{-6}$ with cosine decay. We train all models with one epoch and set the global batch size to 64. We use DeepSpeed with zero2~\citep{deepspeed_zero} and flash-attention-2~\citep{dao2023flashattention2} to improve training efficiency. To train ORMs, a new special token \texttt{<step>} is added at the end of solutions for ORMs training. This makes a consistent formulation for ORMs and PRMs. For PRMs, \texttt{<step>} is added to separate all reasoning steps and assign labels to corresponding tokens. We use ``$+$'' and ``$-$'' as labels to denote the correctness of each step. We use $8\times$ AMD-MI250 GPUs for training and data synthesis. 

\textbf{Results.}  Table~\ref{tab:exp_bon} demonstrates that \ours-ORM and \ours-PRM generally enhance the reasoning performance of MLLMs across policy models of different sizes and benchmarks. Notably, \ours-PRM achieves a \textbf{+10.2} points improvement on the WeMath dataset with Qwen2.5-VL-7B~\citep{qwen2.5vl} as the policy model, compared to the zero-shot baseline. ORMs also exhibit general improvement across all benchmarks and policy models, although some gains are limited, such as +0.4 points on MathVerse with Qwen2.5-VL-72B~\citep{qwen2.5vl}. In contrast, PRMs consistently show substantial performance gains over ORMs, illustrating the benefits of fine-grained rewards in test-time scaling.

Table~\ref{tab:exp_bon_text} highlights that \ours-PRM enhances reasoning abilities for text-only inputs in both the Qwen2.5 series and Ministral-8B models. Specifically, \ours-PRM achieves \textbf{+10.9 points} improvement for Ministral-8B on the challenging MATH~\citep{hendrycks_math} benchmark, demonstrating its effectiveness in text-only scenarios.
\begin{table}[t]
\centering
\caption{\textbf{Results on VisualProcessBench}. We report the F1-score on different datasets and the macro F1 score in overall. We highlight the \textcolor{red2}{\textbf{best}} result across all models.}
\resizebox{\textwidth}{!}{
\begin{tabular}{lcccccc}
\toprule
   & \textbf{MMMU} & \textbf{MathVision} & \textbf{MathVerse} & \textbf{DynaMath} & \textbf{WeMath} & \textbf{Overall} \\ 
\midrule
Random Guessing           & 50.0          & 50.0                & 50.0                  & 50.0              & 50.0            & 50.0             \\ 
\midrule
\multicolumn{7}{c}{\textit{Proprietary Models as Judge}}                                                                                                  \\ \midrule
GPT-4o-mini~\citep{gpt4o}    & 53.6   & 58.9   & 57.1  & 56.7 & 58.5 & 57.9   \\
GPT-4o~\citep{gpt4o}   & 56.3  & 60.2    & 59.7  & 59.0 & 63.3  & 60.3 \\
Gemini-2.0-Flash~\citep{team2023gemini}   & 58.5   & 60.1   & \textcolor{red2}{\textbf{62.8}}  & 66.7  & 58.7  & 62.3  \\ 
\midrule
\multicolumn{7}{c}{\textit{Open-source Models as Judge}}                                                                                                  \\ \midrule
MiniCPM-V2.6-8B~\citep{yao2024minicpm}  & 44.9  & 50.9  & 58.9  & 46.7   & 57.4   & 50.4  \\
LLaVA-OV-7B~\citep{li2025llavaonevision}  & 45.7   & 43.0    & 42.2    & 44.7   & 52.5   & 44.4      \\
LLaVA-OV-72B~\citep{li2025llavaonevision}  & 46.1          & 48.4     & 53.0      & 57.0   & 57.3   & 52.3   \\
Qwen2.5-VL-7B~\citep{qwen2.5vl}    & 53.1      & 51.8      & 47.8       & 51.3   & 54.2     & 51.0   \\
Qwen2.5-VL-72B~\citep{qwen2.5vl}   & 59.2    & 59.0      & 59.7   & 62.9   & 62.3     & 60.5     \\
InternVL2.5-8B~\citep{intern2.5vl}  & 47.1      & 45.5      & 47.8     & 50.3    & 50.8    & 48.0   \\
InternVL2.5-26B~\citep{intern2.5vl}    & 48.8     & 47.4     & 49.2     & 50.4    & 51.4      & 49.2   \\
InternVL2.5-38B~\citep{intern2.5vl}    & 51.5    & 48.4     & 50.9     & 51.8    & 52.5     & 50.8   \\
InternVL2.5-78B~\citep{intern2.5vl}  & 52.0     & 51.7   & 53.7    & 50.8     & 52.5     & 52.6     \\ 
\midrule
\multicolumn{7}{c}{\textit{Multimodal PRMs}} \\
\midrule
VisualPRM-8B~\citep{wang2025visualprm} & 58.5 & \textcolor{red2}{\textbf{62.1}}   & 61.0    & 62.7  & 61.8  & 62.0  \\
\rowcolor{mygray} \ours-PRM & \textcolor{red2}{\textbf{74.1}}& 61.2 & 60.7 & \textcolor{red2}{\textbf{72.7}} &\textcolor{red2}{\textbf{73.8}}  & \textcolor{red2}{\textbf{65.9}}\\
\bottomrule
\end{tabular}
}
\label{tab:exp_vlp}
\end{table}

\subsection{Evaluation on Direct Judgment of Steps}
\label{sec:exp_vlp}
\textbf{Setup.}  In addition to Best-of-N evaluation, we assess \ours-PRM on VisualProcessBench~\citep{wang2025visualprm}, comparing it with general MLLMs and VisualPRM-8B~\citep{wang2025visualprm}. VisualProcessBench~\citep{wang2025visualprm} aims to evaluate the ability to directly judge the correctness of each step. For MLLMs, we prompt the model to analyze and judge the correctness of each step. We report the F1-score for every subset and the micro-F1 score overall.

\textbf{Results.} The results of VisualProcessBench are presented in Table~\ref{tab:exp_vlp}. It is evident that current general open-source MLLMs struggle to judge the correctness of reasoning steps, with most $\app$7B models performing no better than the random guessing baseline. Conversely, multimodal process reward models achieve superior performance, rivaling proprietary models. Notably, our process reward model \ours-PRM, with only 7B parameters, achieves a \textbf{+3.9} points improvement compared to VisualPRM-8B~\citep{wang2025visualprm}, even outperforming proprietary models as judges.

\subsection{Evaluation on the Fine-tuned Model}
\label{sec:exp_finetune}
\begin{table}[t]
\centering
\caption{\textbf{Comparison of multimodal reasoning and mathematical performance across different models}. $\Delta$ denotes improvements compared to Qwen2.5-VL-7B~\citep{qwen2.5}. We highlight improvements at least +0.5 points with \textcolor{darkergreen}{\textbf{green}}.}
\resizebox{0.99\textwidth}{!}{
\begin{tabular}{l*{7}{c}}
\toprule
  & \textbf{MMMU} & \textbf{WeMath} & \textbf{MathVista} & \textbf{MathVision} & \textbf{MathVerse} & \textbf{DynaMath}  & \textbf{LogicVista}  \\
\midrule
\multicolumn{8}{c}{\textit{Proprietary Models}} \\
\midrule
GPT-4o~\citep{gpt4o} &72.9 & 50.6 & 71.6 &43.8 & 49.9 & 48.5 & 64.4 \\
Claude-3.7-Sonnet~\citep{claude3.7} & 71.0 & 49.3 &66.8 &41.9 &46.7 & 39.7 & 58.2 \\	
Gemini-2.0-Pro~\citep{team2023gemini} & 72.6 & 56.5 & 71.3& 48.1 & 67.3 & 43.3 & 53.2 \\
\midrule
\multicolumn{8}{c}{\textit{Open-source Models (>70B)}} \\ 
\midrule
InternVL2.5-78B-MPO~\citep{internvl_mpo} & 68.2 & 37.6 & 76.6 & 36.2 & 43.7 &21.2 & 50.8 \\
InternVL2.5-78B~\citep{intern2.5vl} & 70.1 & 39.8 & 70.6 & 32.2 & 39.2 & 19.2 & 49.0\\
Qwen2.5-VL-72B~\citep{qwen2.5vl} & 68.2 & 49.1 & 74.2 & 39.3 & 47.3 & 35.9 & 55.7 \\
\midrule
\multicolumn{8}{c}{\textit{Open-source Models (\textasciitilde 7B)}} \\ 
\midrule
InternVL2.5-8B~\citep{intern2.5vl} & 56.2 & 20.2 &58.3 & 20.0 & 20.4 &9.2 &33.6 \\
MiniCPM-o-2.6-8B~\citep{yao2024minicpm} & 50.9 & 25.2 & 73.3 & 21.7 & 35.0 & 10.4 & 36.0 \\
Qwen2.5-VL-7B~\citep{qwen2.5vl} & 58.0 & 36.2 & 68.1 & 25.4 & 41.1 & 21.8 & 47.9 \\
\rowcolor{mygray} \ours-7B &  61.1   &  43.0  & 71.4 &  25.7 & 45.7 & 21.9 & 51.3\\
\rowcolor{mygray} $\Delta$ &  \textcolor{darkergreen}{\textbf{+3.1}} & \textcolor{darkergreen}{\textbf{+6.8}} & \textcolor{darkergreen}{\textbf{+3.3}} & +0.3 & \textcolor{darkergreen}{\textbf{+4.6}} & +0.1 & \textcolor{darkergreen}{\textbf{+3.4}}\\
\bottomrule
\end{tabular}
}
\label{tab:exp_benchmark}
\end{table}
\textbf{Setup.} We select Qwen2.5-VL-7B~\citep{qwen2.5vl} as the initial policy model, obtaining training data as outlined in Sec.~\ref{sec:method_application}. We use AdamW~\citep{adamw} optimizer and set the learning rate as $1e^{-6}$ with cosine decay. The batch size is set to 128, and the number of epochs is set to one. We denote our fine-tuned model as \ours-7B and evaluate it on the same benchmarks as Sec.~\ref{sec:bon_exp} in the zero-shot setting. For evaluation, we use VLMEvalKit\footnote{\url{https://github.com/open-compass/VLMEvalKit}}~\citep{duan2024vlmevalkit} 
to evaluate \ours-7B and report the results of all baselines from 
Open LMM Reasoning Leaderboard\footnote{\url{https://huggingface.co/spaces/opencompass/Open_LMM_Reasoning_Leaderboard}}.

\textbf{Results.} ~ As shown in Table~\ref{tab:exp_benchmark}, \ours-7B consistently improves upon Qwen2.5-VL-7B~\citep{qwen2.5vl} across all benchmarks, with significant enhancements of at least $+$0.5 points on five benchmarks. \ours-7B outperforms large open-source models (e.g. InternVL2.5-78B~\citep{intern2.5vl}) on five benchmarks. On the MathVista ~\citep{lu2024mathvista}, \ours-7B even achieves the comparable performance with proprietary models such as Gemini-2.0-Pro~\citep{team2023gemini}, Claude-3.7-Sonnet~\citep{claude3.7}, and GPT-4o~\citep{gpt4o}.

\subsection{Analysis}
\label{sec:exp_analysis}
\begin{figure}[t]
    \centering    
    \begin{subfigure}{0.33\linewidth}
        \includegraphics[width=\linewidth]{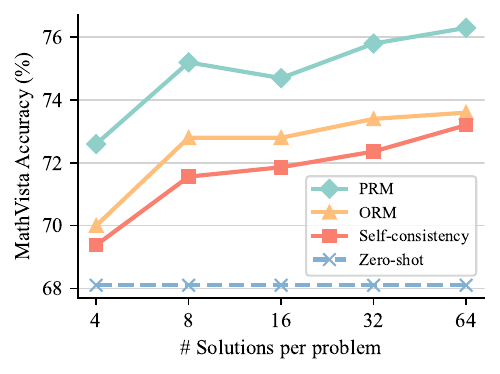}
        \caption{Qwen2.5-VL-7B}
        \label{fig:tts_qwen2.5_7b} 
    \end{subfigure}
    \hfill
    \hspace{-3mm}
    \begin{subfigure}{0.33\linewidth}
        \includegraphics[width=\linewidth]{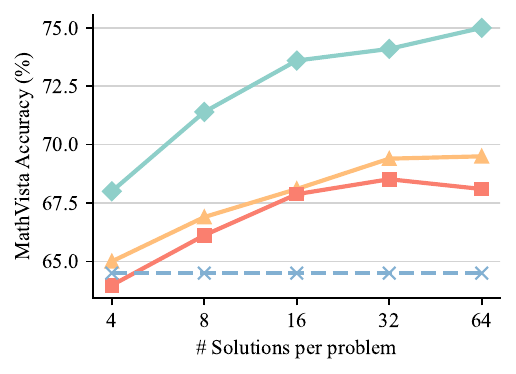}
        \caption{InternVL2.5-8B}
        \label{fig:tts_internvl_8b} 
    \end{subfigure}
    \hfill
    \hspace{-3mm}
    \begin{subfigure}{0.33\linewidth}
        \includegraphics[width=\linewidth]{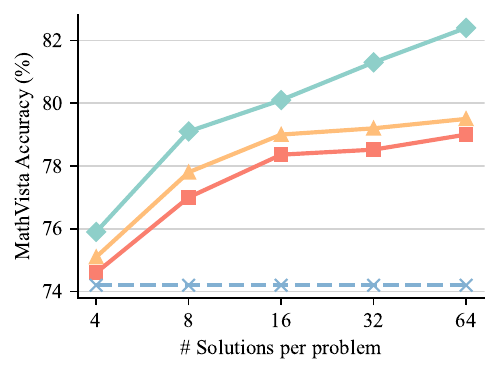}
        \caption{Qwen2.5-VL-72B}
        \label{fig:tts_qwen2.5_72b} 
    \end{subfigure}   
    \caption{Best-of-N results on the MathVista~\citep{lu2024mathvista} across different policies. The number of solutions we sample is from 4 to 64.}
    \label{fig:tts_results}
\end{figure}

\textbf{Test-time scaling.}  To evaluate the scalability of \ours-ORM and \ours-PRM, we compare their performance using different numbers of samples ranging from 4 to 64 on the MathVista~\citep{lu2024mathvista}. As shown in Figure~\ref{fig:tts_results}, reward ranking for solutions at test-time improves reasoning performance across different policy models and number of samples. Specifically, ORMs and PRMs improve the self-consistency~\citep{wang2023selfconsistency} baseline by 0.8 and 2.1 points with Qwen2.5-VL-72B~\citep{qwen2.5vl} using eight samples, respectively. As the sample size increases, the performance improves steadily. However, the performance gain of ORMs is markedly lower than that of PRMs. For instance, \ours-PRM achieves 82.4\% with Qwen2.5-VL-72B~\citep{qwen2.5vl} using 64 samples, while \ours-ORM achieves similar performance with self-consistency~\citep{wang2023selfconsistency} baseline. These findings underscore the advancements and scalability of PRMs in test-time scaling scenarios.
\begin{table}[t]
\centering
\caption{\textbf{Ablation study} of our design choices on the MathVista~\citep{mathvision} and WeMath~\citep{wemath} with Qwen2.5-VL-7B~\citep{qwen2.5vl} as the policy model.}
    \begin{tabular}{lccccc}
    \toprule
    \textbf{Method}      & \textbf{Data} & \textbf{ORM init} & \textbf{Up-sample} & \textbf{MathVista} & \textbf{WeMath} \\ 
    \midrule
    Zero-shot & -    & -     &      -     & 68.1 & 36.2\\
    Self-consistency   & -    & -     & -         & 71.6 & 44.7\\
    \ours-ORM   & \ours-600K & -     & -         & 72.8 & 45.1\\
    \ours-PRM & Random 5K  & \no     & \no       & 73.1 & 45.2 \\
        & \ours-5K    & \no     & \no       & 74.1 & 45.8 \\
        & Random 5K  & \yes     & \no      & 73.6 & 45.4\\
        & \ours-5K    & \yes    & \no       & 74.8 & 46.2\\
        & \ours-5K    & \yes     & \yes    & \textbf{75.2} & \textbf{46.4} \\
        \midrule
        & \textcolor{gray!50}{MC-300K} &  \yes     & \yes     & \textcolor{gray!50}{75.5} & \textcolor{gray!50}{46.4} \\
    \bottomrule
    \end{tabular}
\label{tab:ablation_design}
\end{table}

\begin{figure}
    \centering
    \includegraphics[width=0.7\textwidth]{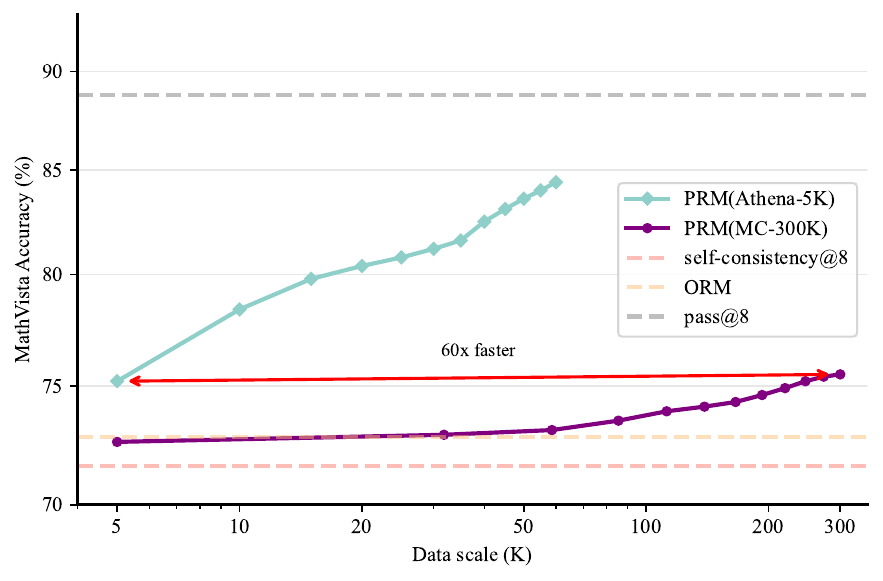}
    \caption{
    Results of data scaling from 5K to 60K. The number of sampled solutions per question is set to 8.
    }
    \label{fig:data_scale}
\end{figure}

\textbf{The effectiveness of our design choices.}~  Table~\ref{tab:ablation_design} presents results under various design choices. Our findings indicate that PRMs trained with selectively chosen 5K samples significantly outperform those trained with randomly selected samples, achieving scores 74.1 \vs 73.1 and 74.8 \vs 73.6 when initialized from ORMs. Moreover, a large-scale dataset of 300K samples labeled using vanilla MC-based methods yields similar performance to our selected 5K samples. With these selected samples, ORM initialization enhances performance by 0.7 points (74.8 \vs 74.1), and up-sampling negative labels further boosts performance by 0.4 points on the MathVista~\citep{lu2024mathvista}.

\textbf{Computation cost analysis.}~ Vanilla MC-based estimation~\citep{wang2024mathshepherd} incurs substantial computational demands to estimate the correctness of intermediate steps. Our prediction consistency filter strategy significantly reduces this burden, requiring only 5K samples for training, thereby lowering the computational cost by a factor of 60. For data synthesis, we employ vLLM primarily~\citep{kwon2023efficient} to accelerate MLLM inference. Compared to the MC-300K dataset, our \ours-5K dataset reduces GPU hours by approximately 45-fold while achieving similar performance.

\textbf{Data scaling}.~ To show the scaling ability of the proposed consistency filter approach, we scale data from 5K to 60K annotated by consistency filter. We add self-consistency, ORM, PRM trained with MC-300K and Pass (Pass means the question is addressed if at last one solution is correct, which is the upper bound in the Best-of-N evaluation.) as baselines, and results are shown in Figure~\ref{fig:data_scale}. From Figure~\ref{fig:data_scale}, we could see that the performance is growing until 60K samples. Although the model trained with 5K samples is slightly worse than the model trained with MC-300K, scaling data to 10K outperforms. Results show that the proposed consistency filter approach is scalable.

\section{Conclusion}
In this paper, we introduce \ours-PRM, trained on 5K high-quality data with process labels. To automatically obtain accurate process labels, we employ a data filtering method based on consistency between weak and strong completers. This strategy significantly reduces computational costs for training and data synthesis while ensuring precise process labels for training PRMs. Additionally, we present two effective strategies for training PRMs: ORM initialization and negative data up-sampling. \ours-PRM demonstrates substantial performance improvements in Best-of-N evaluations, achieving state-of-the-art results on VisualProcessBench. Leveraging \ours-PRM, we train \ours-7B, fine-tuned from Qwen2.5-VL-7B using a reward ranked fine-tuning approach, markedly enhancing the reasoning capabilities of Qwen2.5-VL-7B. Extensive experiments confirm the effectiveness of our proposed methods.

\clearpage
\bibliography{main}

@article{intern2.5vl,
  title={Expanding performance boundaries of open-source multimodal models with model, data, and test-time scaling},
  author={Chen, Zhe and Wang, Weiyun and Cao, Yue and Liu, Yangzhou and Gao, Zhangwei and Cui, Erfei and Zhu, Jinguo and Ye, Shenglong and Tian, Hao and Liu, Zhaoyang and others},
  journal={arXiv preprint arXiv:2412.05271},
  year={2024}
}

@article{qwen2.5vl,
  title={Qwen2. 5-vl technical report},
  author={Bai, Shuai and Chen, Keqin and Liu, Xuejing and Wang, Jialin and Ge, Wenbin and Song, Sibo and Dang, Kai and Wang, Peng and Wang, Shijie and Tang, Jun and others},
  journal={arXiv preprint arXiv:2502.13923},
  year={2025}
}

@inproceedings{zhang2024mathverse,
  title={Mathverse: Does your multi-modal llm truly see the diagrams in visual math problems?},
  author={Zhang, Renrui and Jiang, Dongzhi and Zhang, Yichi and Lin, Haokun and Guo, Ziyu and Qiu, Pengshuo and Zhou, Aojun and Lu, Pan and Chang, Kai-Wei and Qiao, Yu and others},
  booktitle={ECCV},
  year={2024},
}

@inproceedings{wemath,
    title = "We-Math: Does Your Large Multimodal Model Achieve Human-like Mathematical Reasoning?",
    author = "Qiao, Runqi  and
      Tan, Qiuna  and
      Dong, Guanting  and
      MinhuiWu, MinhuiWu  and
      Sun, Chong  and
      Song, Xiaoshuai  and
      Wang, Jiapeng  and
      GongQue, Zhuoma  and
      Lei, Shanglin  and
      Zhang, YiFan  and
      Wei, Zhe  and
      Zhang, Miaoxuan  and
      Qiao, Runfeng  and
      Zong, Xiao  and
      Xu, Yida  and
      Yang, Peiqing  and
      Bao, Zhimin  and
      Diao, Muxi  and
      Li, Chen  and
      Zhang, Honggang",
    booktitle = {ACL},
    year = {2025},
}

@inproceedings{mathvision,
  title={Measuring multimodal mathematical reasoning with math-vision dataset},
  author={Wang, Ke and Pan, Junting and Shi, Weikang and Lu, Zimu and Ren, Houxing and Zhou, Aojun and Zhan, Mingjie and Li, Hongsheng},
    booktitle={NeurIPS},
  year={2024}
}

@inproceedings{lu2024mathvista,
  title={MathVista: Evaluating Mathematical Reasoning of Foundation Models in Visual Contexts},
  author={Lu, Pan and Bansal, Hritik and Xia, Tony and Liu, Jiacheng and Li, Chunyuan and Hajishirzi, Hannaneh and Cheng, Hao and Chang, Kai-Wei and Galley, Michel and Gao, Jianfeng},
  booktitle={ICLR},
  year={2024}
}

@inproceedings{
wang2023selfconsistency,
title={Self-Consistency Improves Chain of Thought Reasoning in Language Models},
author={Xuezhi Wang and Jason Wei and Dale Schuurmans and Quoc V Le and Ed H. Chi and Sharan Narang and Aakanksha Chowdhery and Denny Zhou},
booktitle={ICLR},
year={2023},
}

@inproceedings{
lightman2024lets,
title={Let's Verify Step by Step},
author={Hunter Lightman and Vineet Kosaraju and Yuri Burda and Harrison Edwards and Bowen Baker and Teddy Lee and Jan Leike and John Schulman and Ilya Sutskever and Karl Cobbe},
booktitle={ICLR},
year={2024},
}

@article{cobbe2021training,
  title={Training verifiers to solve math word problems},
  author={Cobbe, Karl and Kosaraju, Vineet and Bavarian, Mohammad and Chen, Mark and Jun, Heewoo and Kaiser, Lukasz and Plappert, Matthias and Tworek, Jerry and Hilton, Jacob and Nakano, Reiichiro and others},
  journal={arXiv preprint arXiv:2110.14168},
  year={2021}
}

@inproceedings{shi2024math,
  title={Math-LLaVA: Bootstrapping Mathematical Reasoning for Multimodal Large Language Models},
  author={Shi, Wenhao and Hu, Zhiqiang and Bin, Yi and Liu, Junhua and Yang, Yang and Ng, See Kiong and Bing, Lidong and Lee, Roy},
  booktitle={EMNLP},
  year={2024}
}

@inproceedings{wang2024mathshepherd,
  title={Math-Shepherd: Verify and Reinforce {LLM}s Step-by-step without Human Annotations},
  author={Wang, Peiyi and Li, Lei and Shao, Zhihong and Xu, Runxin and Dai, Damai and Li, Yifei and Chen, Deli and Wu, Yu and Sui, Zhifang},
  booktitle={ACL},
  year={2024}
}

@inproceedings{hendrycks_math,
 author = {Hendrycks, Dan and Burns, Collin and Kadavath, Saurav and Arora, Akul and Basart, Steven and Tang, Eric and Song, Dawn and Steinhardt, Jacob},
 booktitle = {NeurIPS},
 title = {Measuring Mathematical Problem Solving With the MATH Dataset},
 year = {2021}
}

@inproceedings{llava,
 author = {Liu, Haotian and Li, Chunyuan and Wu, Qingyang and Lee, Yong Jae},
 booktitle = {NeurIPS},
 title = {Visual Instruction Tuning},
 year = {2023}
}

@inproceedings{
snell2025scaling,
title={Scaling {LLM} Test-Time Compute Optimally Can be More Effective than Scaling Parameters for Reasoning},
author={Charlie Victor Snell and Jaehoon Lee and Kelvin Xu and Aviral Kumar},
booktitle={ICLR},
year={2025},
}

@article{OmegaPRM,
  title={Improve mathematical reasoning in language models by automated process supervision},
  author={Luo, Liangchen and Liu, Yinxiao and Liu, Rosanne and Phatale, Samrat and Guo, Meiqi and Lara, Harsh and Li, Yunxuan and Shu, Lei and Zhu, Yun and Meng, Lei and others},
  journal={arXiv preprint arXiv:2406.06592},
  year={2024}
}

@inproceedings{
setlur2025rewarding,
title={Rewarding Progress: Scaling Automated Process Verifiers for {LLM} Reasoning},
author={Amrith Setlur and Chirag Nagpal and Adam Fisch and Xinyang Geng and Jacob Eisenstein and Rishabh Agarwal and Alekh Agarwal and Jonathan Berant and Aviral Kumar},
booktitle={ICLR},
year={2025},
}

@inproceedings{
cot,
title={Chain of Thought Prompting Elicits Reasoning in Large Language Models},
author={Jason Wei and Xuezhi Wang and Dale Schuurmans and Maarten Bosma and brian ichter and Fei Xia and Ed H. Chi and Quoc V Le and Denny Zhou},
booktitle={NeurIPS},
year={2022},
}

@article{
rlhf_workflow,
title={{RLHF} Workflow: From Reward Modeling to Online {RLHF}},
author={Hanze Dong and Wei Xiong and Bo Pang and Haoxiang Wang and Han Zhao and Yingbo Zhou and Nan Jiang and Doyen Sahoo and Caiming Xiong and Tong Zhang},
journal={Transactions on Machine Learning Research},
year={2024},
}

@inproceedings{duan2024vlmevalkit,
  title={{VLMEvalKit}: An open-source toolkit for evaluating large multi-modality models},
  author={Duan, Haodong and Yang, Junming and Qiao, Yuxuan and Fang, Xinyu and Chen, Lin and Liu, Yuan and Dong, Xiaoyi and Zang, Yuhang and Zhang, Pan and Wang, Jiaqi and others},
  booktitle={ACM MM},
  year={2024}
}

@inproceedings{yue2024mmmu,
  title={{MMMU}: A massive multi-discipline multimodal understanding and reasoning benchmark for expert agi},
  author={Yue, Xiang and Ni, Yuansheng and Zhang, Kai and Zheng, Tianyu and Liu, Ruoqi and Zhang, Ge and Stevens, Samuel and Jiang, Dongfu and Ren, Weiming and Sun, Yuxuan and others},
  booktitle={CVPR},
  year={2024}
}

@article{qwen2.5,
  title={Qwen2. 5 technical report},
  author={Yang, An and Yang, Baosong and Zhang, Beichen and Hui, Binyuan and Zheng, Bo and Yu, Bowen and Li, Chengyuan and Liu, Dayiheng and Huang, Fei and Wei, Haoran and others},
  journal={arXiv preprint arXiv:2412.15115},
  year={2024}
}

@article{wang2025visualprm,
  title={Visual{PRM}: An effective process reward model for multimodal reasoning},
  author={Wang, Weiyun and Gao, Zhangwei and Chen, Lianjie and Chen, Zhe and Zhu, Jinguo and Zhao, Xiangyu and Liu, Yangzhou and Cao, Yue and Ye, Shenglong and Zhu, Xizhou and others},
  journal={arXiv preprint arXiv:2503.10291},
  year={2025}
}

@inproceedings{gpt3,
 author = {Brown, Tom and Mann, Benjamin and Ryder, Nick and Subbiah, Melanie and Kaplan, Jared D and Dhariwal, Prafulla and Neelakantan, Arvind and Shyam, Pranav and Sastry, Girish and Askell, Amanda and Agarwal, Sandhini and Herbert-Voss, Ariel and Krueger, Gretchen and Henighan, Tom and Child, Rewon and Ramesh, Aditya and Ziegler, Daniel and Wu, Jeffrey and Winter, Clemens and Hesse, Chris and Chen, Mark and Sigler, Eric and Litwin, Mateusz and Gray, Scott and Chess, Benjamin and Clark, Jack and Berner, Christopher and McCandlish, Sam and Radford, Alec and Sutskever, Ilya and Amodei, Dario},
 booktitle = {NeurIPS},
 title = {Language Models are Few-Shot Learners},
 year = {2020}
}

@inproceedings{chartqa,
    title = "{C}hart{QA}: A Benchmark for Question Answering about Charts with Visual and Logical Reasoning",
    author = {Masry, Ahmed  and
      Long, Do  and
      Tan, Jia Qing  and
      Joty, Shafiq  and
      Hoque, Enamul},
    booktitle = {ACL},
    year = {2022},
}

@article{
dong2023raft,
title={{RAFT}: Reward rAnked FineTuning for Generative Foundation Model Alignment},
author={Hanze Dong and Wei Xiong and Deepanshu Goyal and Yihan Zhang and Winnie Chow and Rui Pan and Shizhe Diao and Jipeng Zhang and KaShun SHUM and Tong Zhang},
journal={Transactions on Machine Learning Research},
year={2023},
}

@article{yuan2023scaling,
  title={Scaling relationship on learning mathematical reasoning with large language models},
  author={Yuan, Zheng and Yuan, Hongyi and Li, Chengpeng and Dong, Guanting and Lu, Keming and Tan, Chuanqi and Zhou, Chang and Zhou, Jingren},
  journal={arXiv preprint arXiv:2308.01825},
  year={2023}
}

@inproceedings{deepspeed_zero,
  author       = {Samyam Rajbhandari and
                  Jeff Rasley and
                  Olatunji Ruwase and
                  Yuxiong He},
  title        = {ZeRO: memory optimizations toward training trillion parameter models},
  booktitle    = {Proceedings of the International Conference for High Performance Computing,
                  Networking, Storage and Analysis},
  year         = {2020},
}

@article{geomverse,
  title={Geomverse: A systematic evaluation of large models for geometric reasoning},
  author={Kazemi, Mehran and Alvari, Hamidreza and Anand, Ankit and Wu, Jialin and Chen, Xi and Soricut, Radu},
  journal={arXiv preprint arXiv:2312.12241},
  year={2023}
}

@article{chen2022unigeo,
  title={Unigeo: Unifying geometry logical reasoning via reformulating mathematical expression},
  author={Chen, Jiaqi and Li, Tong and Qin, Jinghui and Lu, Pan and Lin, Liang and Chen, Chongyu and Liang, Xiaodan},
  journal={arXiv preprint arXiv:2212.02746},
  year={2022}
}

@article{geo3k,
  title={Inter-GPS: Interpretable geometry problem solving with formal language and symbolic reasoning},
  author={Lu, Pan and Gong, Ran and Jiang, Shibiao and Qiu, Liang and Huang, Siyuan and Liang, Xiaodan and Zhu, Song-Chun},
  journal={arXiv preprint arXiv:2105.04165},
  year={2021}
}

@inproceedings{
zou2025dynamath,
title={DynaMath: A Dynamic Visual Benchmark for Evaluating Mathematical Reasoning Robustness of Vision Language Models},
author={Chengke Zou and Xingang Guo and Rui Yang and Junyu Zhang and Bin Hu and Huan Zhang},
booktitle={ICLR},
year={2025},
}

@article{xiao2024logicvista,
  title={Logicvista: Multimodal llm logical reasoning benchmark in visual contexts},
  author={Xiao, Yijia and Sun, Edward and Liu, Tianyu and Wang, Wei},
  journal={arXiv preprint arXiv:2407.04973},
  year={2024}
}

@inproceedings{dao2023flashattention2,
  title={Flash{A}ttention-2: Faster Attention with Better Parallelism and Work Partitioning},
  author={Dao, Tri},
  booktitle={ICLR},
  year={2024}
}

@article{
scale_selftraining,
title={Beyond Human Data: Scaling Self-Training for Problem-Solving with Language Models},
author={Avi Singh and John D Co-Reyes and Rishabh Agarwal and Ankesh Anand and Piyush Patil and Xavier Garcia and Peter J Liu and James Harrison and Jaehoon Lee and Kelvin Xu and Aaron T Parisi and Abhishek Kumar and Alexander A Alemi and Alex Rizkowsky and Azade Nova and Ben Adlam and Bernd Bohnet and Gamaleldin Fathy Elsayed and Hanie Sedghi and Igor Mordatch and Isabelle Simpson and Izzeddin Gur and Jasper Snoek and Jeffrey Pennington and Jiri Hron and Kathleen Kenealy and Kevin Swersky and Kshiteej Mahajan and Laura A Culp and Lechao Xiao and Maxwell Bileschi and Noah Constant and Roman Novak and Rosanne Liu and Tris Warkentin and Yamini Bansal and Ethan Dyer and Behnam Neyshabur and Jascha Sohl-Dickstein and Noah Fiedel},
journal={Transactions on Machine Learning Research},
year={2024},
}

@inproceedings{
zeng2025bstar,
title={B-{ST}aR: Monitoring and Balancing Exploration and Exploitation in Self-Taught Reasoners},
author={Weihao Zeng and Yuzhen Huang and Lulu Zhao and Yijun Wang and Zifei Shan and Junxian He},
booktitle={ICLR},
year={2025},
}

@article{lindstrom2022clevr,
  title={Clevr-math: A dataset for compositional language, visual and mathematical reasoning},
  author={Lindstr{\"o}m, Adam Dahlgren and Abraham, Savitha Sam},
  journal={arXiv preprint arXiv:2208.05358},
  year={2022}
}

@inproceedings{
scienceqa,
title={Learn to Explain: Multimodal Reasoning via Thought Chains for Science Question Answering},
author={Pan Lu and Swaroop Mishra and Tony Xia and Liang Qiu and Kai-Wei Chang and Song-Chun Zhu and Oyvind Tafjord and Peter Clark and Ashwin Kalyan},
booktitle={NeurIPS},
year={2022},
}

@article{
li2025llavaonevision,
title={{LL}a{VA}-OneVision: Easy Visual Task Transfer},
author={Bo Li and Yuanhan Zhang and Dong Guo and Renrui Zhang and Feng Li and Hao Zhang and Kaichen Zhang and Peiyuan Zhang and Yanwei Li and Ziwei Liu and Chunyuan Li},
journal={Transactions on Machine Learning Research},
year={2025},
}

@inproceedings{
lu2024autopsv,
title={Auto{PSV}: Automated Process-Supervised Verifier},
author={Jianqiao Lu and Zhiyang Dou and Hongru Wang and Zeyu Cao and Jianbo Dai and Yunlong Feng and Zhijiang Guo},
booktitle={NeurIPS},
year={2024},
}

@article{jiang2023mistral7b,
  title={Mistral 7B},
  author={Albert Q. Jiang and Alexandre Sablayrolles and Arthur Mensch and Chris Bamford and Devendra Singh Chaplot and Diego de las Casas and Florian Bressand and Gianna Lengyel and Guillaume Lample and Lucile Saulnier and Lélio Renard Lavaud and Marie-Anne Lachaux and Pierre Stock and Teven Le Scao and Thibaut Lavril and Thomas Wang and Timothée Lacroix and William El Sayed},
  journal={arXiv preprint arXiv:2310.06825},
  year={2023}
}

@misc{gpt4o,
    author = {OpenAI},
    howpublished = {\url{https://openai.com/index/hello-gpt-4o/}},
    note = {Accessed: 2024-02-09, 2024-02-11, 2024-02-12},
    title = {Hello gpt-4o},
    year = {2024}
}

@misc{claude3.7,
    author = {Anthropic},
    note = {Accessed on Feb 25, 2025},
    title = {Claude 3.7 Sonnet and Claude Code},
    url = {https://www.anthropic.com/news/claude-3-7-sonnet},
    year = {2025}
}

@article{team2023gemini,
  title={Gemini: a family of highly capable multimodal models},
  author={{Gemini Team}},
  journal={arXiv preprint arXiv:2312.11805},
  year={2023}
}

@article{uesato2022solving,
  title={Solving math word problems with process-and outcome-based feedback},
  author={Uesato, Jonathan and Kushman, Nate and Kumar, Ramana and Song, Francis and Siegel, Noah and Wang, Lisa and Creswell, Antonia and Irving, Geoffrey and Higgins, Irina},
  journal={arXiv preprint arXiv:2211.14275},
  year={2022}
}

@article{llama2,
  title={Llama 2: Open foundation and fine-tuned chat models},
  author={Touvron, Hugo and Martin, Louis and Stone, Kevin and Albert, Peter and Almahairi, Amjad and Babaei, Yasmine and Bashlykov, Nikolay and Batra, Soumya and Bhargava, Prajjwal and Bhosale, Shruti and others},
  journal={arXiv preprint arXiv:2307.09288},
  year={2023}
}

@article{llama3,
  title={The llama 3 herd of models},
  author={Grattafiori, Aaron and Dubey, Abhimanyu and Jauhri, Abhinav and Pandey, Abhinav and Kadian, Abhishek and Al-Dahle, Ahmad and Letman, Aiesha and Mathur, Akhil and Schelten, Alan and Vaughan, Alex and others},
  journal={arXiv preprint arXiv:2407.21783},
  year={2024}
}

@article{deepseekv3,
  title={Deepseek-v3 technical report},
  author={Liu, Aixin and Feng, Bei and Xue, Bing and Wang, Bingxuan and Wu, Bochao and Lu, Chengda and Zhao, Chenggang and Deng, Chengqi and Zhang, Chenyu and Ruan, Chong and others},
  journal={arXiv preprint arXiv:2412.19437},
  year={2024}
}

@inproceedings{antol2015vqa,
  title={{VQA}: Visual question answering},
  author={Antol, Stanislaw and Agrawal, Aishwarya and Lu, Jiasen and Mitchell, Margaret and Batra, Dhruv and Zitnick, C Lawrence and Parikh, Devi},
  booktitle={ICCV},
  year={2015}
}

@inproceedings{mathew2021docvqa,
  title={DocVQA: {A} Dataset for {VQA} on Document Images},
  author={Mathew, Minesh and Karatzas, Dimosthenis and Jawahar, CV},
  booktitle={WACV},
  year={2021}
}

@article{internvl_mpo,
  title={Enhancing the reasoning ability of multimodal large language models via mixed preference optimization},
  author={Wang, Weiyun and Chen, Zhe and Wang, Wenhai and Cao, Yue and Liu, Yangzhou and Gao, Zhangwei and Zhu, Jinguo and Zhu, Xizhou and Lu, Lewei and Qiao, Yu and others},
  journal={arXiv preprint arXiv:2411.10442},
  year={2024}
}

@inproceedings{iterative_reasoning_dpo,
title={Iterative Reasoning Preference Optimization},
author={Richard Yuanzhe Pang and Weizhe Yuan and He He and Kyunghyun Cho and Sainbayar Sukhbaatar and Jason E Weston},
booktitle={NeurIPS},
year={2024},
}

@article{shao2024deepseekmath,
  title={Deepseekmath: Pushing the limits of mathematical reasoning in open language models},
  author={Shao, Zhihong and Wang, Peiyi and Zhu, Qihao and Xu, Runxin and Song, Junxiao and Bi, Xiao and Zhang, Haowei and Zhang, Mingchuan and Li, YK and Wu, Y and others},
  journal={arXiv preprint arXiv:2402.03300},
  year={2024}
}

@article{deepseekr1,
  title = {DeepSeek-R1 incentivizes reasoning in LLMs through reinforcement learning},
  volume = {645},
  number = {8081},
  journal = {Nature},
  publisher = {Springer Science and Business Media LLC},
  author = {Guo,  Daya and Yang,  Dejian and Zhang,  Haowei and Song,  Junxiao and Wang,  Peiyi and Zhu,  Qihao and Xu,  Runxin and Zhang,  Ruoyu and Ma,  Shirong and Bi,  Xiao and Zhang,  Xiaokang and Yu,  Xingkai and Wu,  Yu and Wu,  Z. F. and Gou,  Zhibin and Shao,  Zhihong and Li,  Zhuoshu and Gao,  Ziyi and Liu,  Aixin and Xue,  Bing and Wang,  Bingxuan and Wu,  Bochao and Feng,  Bei and Lu,  Chengda and Zhao,  Chenggang and Deng,  Chengqi and Ruan,  Chong and Dai,  Damai and Chen,  Deli and Ji,  Dongjie and Li,  Erhang and Lin,  Fangyun and Dai,  Fucong and Luo,  Fuli and Hao,  Guangbo and Chen,  Guanting and Li,  Guowei and Zhang,  H. and Xu,  Hanwei and Ding,  Honghui and Gao,  Huazuo and Qu,  Hui and Li,  Hui and Guo,  Jianzhong and Li,  Jiashi and Chen,  Jingchang and Yuan,  Jingyang and Tu,  Jinhao and Qiu,  Junjie and Li,  Junlong and Cai,  J. L. and Ni,  Jiaqi and Liang,  Jian and Chen,  Jin and Dong,  Kai and Hu,  Kai and You,  Kaichao and Gao,  Kaige and Guan,  Kang and Huang,  Kexin and Yu,  Kuai and Wang,  Lean and Zhang,  Lecong and Zhao,  Liang and Wang,  Litong and Zhang,  Liyue and Xu,  Lei and Xia,  Leyi and Zhang,  Mingchuan and Zhang,  Minghua and Tang,  Minghui and Zhou,  Mingxu and Li,  Meng and Wang,  Miaojun and Li,  Mingming and Tian,  Ning and Huang,  Panpan and Zhang,  Peng and Wang,  Qiancheng and Chen,  Qinyu and Du,  Qiushi and Ge,  Ruiqi and Zhang,  Ruisong and Pan,  Ruizhe and Wang,  Runji and Chen,  R. J. and Jin,  R. L. and Chen,  Ruyi and Lu,  Shanghao and Zhou,  Shangyan and Chen,  Shanhuang and Ye,  Shengfeng and Wang,  Shiyu and Yu,  Shuiping and Zhou,  Shunfeng and Pan,  Shuting and Li,  S. S. and Zhou,  Shuang and Wu,  Shaoqing and Yun,  Tao and Pei,  Tian and Sun,  Tianyu and Wang,  T. and Zeng,  Wangding and Liu,  Wen and Liang,  Wenfeng and Gao,  Wenjun and Yu,  Wenqin and Zhang,  Wentao and Xiao,  W. L. and An,  Wei and Liu,  Xiaodong and Wang,  Xiaohan and Chen,  Xiaokang and Nie,  Xiaotao and Cheng,  Xin and Liu,  Xin and Xie,  Xin and Liu,  Xingchao and Yang,  Xinyu and Li,  Xinyuan and Su,  Xuecheng and Lin,  Xuheng and Li,  X. Q. and Jin,  Xiangyue and Shen,  Xiaojin and Chen,  Xiaosha and Sun,  Xiaowen and Wang,  Xiaoxiang and Song,  Xinnan and Zhou,  Xinyi and Wang,  Xianzu and Shan,  Xinxia and Li,  Y. K. and Wang,  Y. Q. and Wei,  Y. X. and Zhang,  Yang and Xu,  Yanhong and Li,  Yao and Zhao,  Yao and Sun,  Yaofeng and Wang,  Yaohui and Yu,  Yi and Zhang,  Yichao and Shi,  Yifan and Xiong,  Yiliang and He,  Ying and Piao,  Yishi and Wang,  Yisong and Tan,  Yixuan and Ma,  Yiyang and Liu,  Yiyuan and Guo,  Yongqiang and Ou,  Yuan and Wang,  Yuduan and Gong,  Yue and Zou,  Yuheng and He,  Yujia and Xiong,  Yunfan and Luo,  Yuxiang and You,  Yuxiang and Liu,  Yuxuan and Zhou,  Yuyang and Zhu,  Y. X. and Huang,  Yanping and Li,  Yaohui and Zheng,  Yi and Zhu,  Yuchen and Ma,  Yunxian and Tang,  Ying and Zha,  Yukun and Yan,  Yuting and Ren,  Z. Z. and Ren,  Zehui and Sha,  Zhangli and Fu,  Zhe and Xu,  Zhean and Xie,  Zhenda and Zhang,  Zhengyan and Hao,  Zhewen and Ma,  Zhicheng and Yan,  Zhigang and Wu,  Zhiyu and Gu,  Zihui and Zhu,  Zijia and Liu,  Zijun and Li,  Zilin and Xie,  Ziwei and Song,  Ziyang and Pan,  Zizheng and Huang,  Zhen and Xu,  Zhipeng and Zhang,  Zhongyu and Zhang,  Zhen},
  year = {2025},
  month = sep,
  pages = {633–638}
}

@article{hu2025reinforce++,
  title={REINFORCE++: A Simple and Efficient Approach for Aligning Large Language Models},
  author={Hu, Jian},
  journal={arXiv preprint arXiv:2501.03262},
  year={2025}
}

@article{team2025kimi,
  title={Kimi k1. 5: Scaling reinforcement learning with llms},
  author={Team, Kimi and Du, Angang and Gao, Bofei and Xing, Bowei and Jiang, Changjiu and Chen, Cheng and Li, Cheng and Xiao, Chenjun and Du, Chenzhuang and Liao, Chonghua and others},
  journal={arXiv preprint arXiv:2501.12599},
  year={2025}
}

@inproceedings{muennighoff2025s1,
    title = "s1: Simple test-time scaling",
    author = "Muennighoff, Niklas  and
      Yang, Zitong  and
      Shi, Weijia  and
      Li, Xiang Lisa  and
      Fei-Fei, Li  and
      Hajishirzi, Hannaneh  and
      Zettlemoyer, Luke  and
      Liang, Percy  and
      Candes, Emmanuel  and
      Hashimoto, Tatsunori",
    booktitle = "EMNLP",
    year = "2025",
}

@article{yao2024minicpm,
  title={Efficient GPT-4V level multimodal large language model for deployment on edge devices},
  author={Yao, Yuan and Yu, Tianyu and Zhang, Ao and Wang, Chongyi and Cui, Junbo and Zhu, Hongji and Cai, Tianchi and Chen, Chi and Li, Haoyu and Zhao, Weilin and others},
  journal={Nature Communications},
  volume={16},
  number={1},
  pages={5509},
  year={2025},
}

@inproceedings{
yuan2024implicitprm,
title={Free Process Rewards without Process Labels},
author={Lifan Yuan and Wendi Li and Huayu Chen and Ganqu Cui and Ning Ding and Kaiyan Zhang and Bowen Zhou and Zhiyuan Liu and Hao Peng},
booktitle={ICML},
year={2025},
}

@article{cui2025process,
  title={Process reinforcement through implicit rewards},
  author={Cui, Ganqu and Yuan, Lifan and Wang, Zefan and Wang, Hanbin and Li, Wendi and He, Bingxiang and Fan, Yuchen and Yu, Tianyu and Xu, Qixin and Chen, Weize and others},
  journal={arXiv preprint arXiv:2502.01456},
  year={2025}
}

@inproceedings{
li2025process_q_ranking,
title={Process Reward Model with Q-value Rankings},
author={Wendi Li and Yixuan Li},
booktitle={ICLR},
year={2025}
}

@inproceedings{
adamw,
title={Decoupled Weight Decay Regularization},
author={Ilya Loshchilov and Frank Hutter},
booktitle={ICLR},
year={2019},
}

@inproceedings{
top_p,
title={The Curious Case of Neural Text Degeneration},
author={Ari Holtzman and Jan Buys and Li Du and Maxwell Forbes and Yejin Choi},
booktitle={ICLR},
year={2020},
}

@inproceedings{instruct_gpt,
title={Training language models to follow instructions with human feedback},
author={Long Ouyang and Jeffrey Wu and Xu Jiang and Diogo Almeida and Carroll Wainwright and Pamela Mishkin and Chong Zhang and Sandhini Agarwal and Katarina Slama and Alex Gray and John Schulman and Jacob Hilton and Fraser Kelton and Luke Miller and Maddie Simens and Amanda Askell and Peter Welinder and Paul Christiano and Jan Leike and Ryan Lowe},
booktitle={NeurIPS},
year={2022},
}

@article{bai2022training,
  title={Training a helpful and harmless assistant with reinforcement learning from human feedback},
  author={Bai, Yuntao and Jones, Andy and Ndousse, Kamal and Askell, Amanda and Chen, Anna and DasSarma, Nova and Drain, Dawn and Fort, Stanislav and Ganguli, Deep and Henighan, Tom and others},
  journal={arXiv preprint arXiv:2204.05862},
  year={2022}
}

@article{kaplan2020scaling,
  title={Scaling laws for neural language models},
  author={Kaplan, Jared and McCandlish, Sam and Henighan, Tom and Brown, Tom B and Chess, Benjamin and Child, Rewon and Gray, Scott and Radford, Alec and Wu, Jeffrey and Amodei, Dario},
  journal={arXiv preprint arXiv:2001.08361},
  year={2020}
}

@article{hoffmann2022training,
  title={Training compute-optimal large language models},
  author={Hoffmann, Jordan and Borgeaud, Sebastian and Mensch, Arthur and Buchatskaya, Elena and Cai, Trevor and Rutherford, Eliza and Casas, Diego de Las and Hendricks, Lisa Anne and Welbl, Johannes and Clark, Aidan and others},
  journal={arXiv preprint arXiv:2203.15556},
  year={2022}
}

@article{tts-survey,
  title={What, How, Where, and How Well? A Survey on Test-Time Scaling in Large Language Models},
  author={Zhang, Qiyuan and Lyu, Fuyuan and Sun, Zexu and Wang, Lei and Zhang, Weixu and Guo, Zhihan and Wang, Yufei and King, Irwin and Liu, Xue and Ma, Chen},
  journal={arXiv preprint arXiv:2503.24235},
  year={2025}
}

@inproceedings{kwon2023efficient,
  title={Efficient Memory Management for Large Language Model Serving with PagedAttention},
  author={Woosuk Kwon and Zhuohan Li and Siyuan Zhuang and Ying Sheng and Lianmin Zheng and Cody Hao Yu and Joseph E. Gonzalez and Hao Zhang and Ion Stoica},
  booktitle={Proceedings of the ACM SIGOPS 29th Symposium on Operating Systems Principles},
  year={2023}
}

@inproceedings{zhang2025lessons,
    title = "The Lessons of Developing Process Reward Models in Mathematical Reasoning",
    author = "Zhang, Zhenru  and
      Zheng, Chujie  and
      Wu, Yangzhen  and
      Zhang, Beichen  and
      Lin, Runji  and
      Yu, Bowen  and
      Liu, Dayiheng  and
      Zhou, Jingren  and
      Lin, Junyang",
    booktitle = "ACL",
    year = "2025",
}

@inproceedings{cao2022augmented,
  title={An augmented benchmark dataset for geometric question answering through dual parallel text encoding},
  author={Cao, Jie and Xiao, Jing},
  booktitle={Proceedings of the 29th international conference on computational linguistics},
  pages={1511--1520},
  year={2022}
}

@misc{numina_math_datasets,
  author = {Jia LI and Edward Beeching and Lewis Tunstall and Ben Lipkin and Roman Soletskyi and Shengyi Costa Huang and Kashif Rasul and Longhui Yu and Albert Jiang and Ziju Shen and Zihan Qin and Bin Dong and Li Zhou and Yann Fleureau and Guillaume Lample and Stanislas Polu},
  title = {NuminaMath},
  year = {2024},
  url={https://huggingface.co/datasets/AI-MO/NuminaMath-1.5},
}

@article{multi-rm,
  title   = {Rethinking Reward Models for Multi-Domain Test-Time Scaling},
  author  = {Lee, Dong Bok and Lee, Seanie and Park, Sangwoo and Kang, Minki and Baek, Jinheon and Kim, Dongki and Wagner, Dominik and Jin, Jiongdao and Lee, Heejun and Bocklet, Tobias and Wang, Jinyu and Fu, Jingjing and Hwang, Sung Ju and Bian, Jiang and Song, Lei},
  journal = {arXiv preprint arXiv:2510.00492},
  year    = {2025}
}
\bibliographystyle{tmlr}

\clearpage
\appendix

\section{More Experiments and Details}
\subsection{Usage of PRMs}
We show a python-style code snippet to demonstrate how to use the \ours-PRM with \texttt{transformers} to get the reward for each step in Algorithm~\ref{alg:prm_infer_code}.

\begin{algorithm}[t]
\caption{Pseudocode of PRM inference in a PyTorch-like style.}
\label{alg:prm_infer_code}
\definecolor{codeblue}{rgb}{0.25,0.5,0.5}
\lstset{
  backgroundcolor=\color{white},
  basicstyle=\fontsize{7.2pt}{7.2pt}\ttfamily\selectfont,
  columns=fullflexible,
  breaklines=true,
  captionpos=b,
  commentstyle=\fontsize{7.2pt}{7.2pt}\color{codeblue},
  keywordstyle=\fontsize{7.2pt}{7.2pt},
  showstringspaces=false,
}
\begin{lstlisting}[language=python]
question = "What is the highest number shown on the black outer part of the watch?"  # question for test  
response = """To determine the highest number shown on the black outer part of the watch, let's analyze the image step by step.

1. The black outer part of the watch appears to be a bezel with numbers printed on it. 

2. The numbers on the bezel are arranged in a circular pattern. 

3. Observing the numbers, they range from 0 to 60, increasing in increments of 5.

4. The highest number visible on the bezel is 60. 

Therefore, the highest number shown on the black outer part of the watch is 60.

Answer:60
""" # response generated by the policy model

# load model and tokenizer
from transformers import Qwen2_5_VLForConditionalGeneration, AutoTokenizer, AutoProcessor
from qwen_vl_utils import process_vision_info

good_token = "+" # token denotes correct steps
bad_token = "-"  # token denotes incorrect steps
step_token = "<step>"  # token for separate different steps
steps = response.split("\n\n")
response = step_token.join(steps) + step_token

model_path = "Athena-PRM-7B"
model = Qwen2_5_VLForConditionalGeneration.from_pretrained(model_path, torch_dtype="auto", device_map="auto")
processor = AutoProcessor.from_pretrained(model_path)
tokenizer = processor.tokenizer
step_tag_id = tokenizer.encode(f"{step_token}")[0]
good_token_id = tokenizer.encode(f"{good_token}")[0]
bad_token_id = tokenizer.encode(f"{bad_token}")[0]
messages = [
    {
        "role": "user",
        "content": [
            {
                "type": "image",
                "image": "test_image.jpg",  # image path for test
            },
            {"type": "text", "text": question},
            ],
    },
    {
        "role":"assistant",
        "content":steps,
    },
]

# Preparation for inference
text = processor.apply_chat_template(
    messages, tokenize=False, add_generation_prompt=True
)
image_inputs, video_inputs = process_vision_info(messages)
inputs = processor(
    text=[text],
    images=image_inputs,
    videos=video_inputs,
    padding=True,
    return_tensors="pt",
)
inputs = inputs.to("cuda")
logits = model(inputs) # logits shape: [batch size, sequence length, vocabulary size]
logits = logits[:,:,[good_token_id,bad_token_id]]  # only keep logits of +/-
scores = logits.softmax(dim=-1)[:,:,0] # use probability of token + as the step reward
step_scores = scores[input_id == step_tag_id]
print(step_scores) # get the reward of each step
\end{lstlisting}
\end{algorithm}

\subsection{Choices of Completers}
\label{sec:appendix_choice_completer}
Here we verify the effect of combining different completers on the accuracy of estimated labels. Results are listed in Table~\ref{tab:exp_completer}. From Table~\ref{tab:exp_completer}, it is noted that the consistency filter for two completers usually improves the accuracy of estimated labels. For example, combining two weak or strong completers improves accuracy by almost two points in general. Furthermore, the accuracy of estimated labels is improved by combining one weak and one strong completer, achieving the most competitive accuracy, e.g., 95.2\% with weak completer $\phi_{w}^2$ and strong completer $\phi_{s}^1$. This shows that the proposed consistency filter strategy is effective across all combinations of different completers and achieves the best performance when we combine a weak completer and a strong completer. We also explore use more completers to further improve the quality of process labels. However, we did not observe any significant improvement, despite the increased computational cost.

\begin{table}[htbp]
\centering
\caption{The accuracy of estimated process labels with Different completers combination. In Two Completers, we use consistency filter introduced in Sec.~\ref{subsec:auto_prm}. $\phi_{w}^1$, $\phi_{w}^2$, $\phi_{s}^1$, and $\phi_{s}^2$ denote  Mistral-7B-Instruct, Qwen2-7B-Instruct, Qwen2.5-72B-Instruct and Llama-3-70B-Instruct, respectively.}
\begin{tabular}{lccccc}
\toprule
\multicolumn{1}{l}{} & \multicolumn{2}{c}{Weak Completer} & \multicolumn{2}{c}{Strong Completer} & \multicolumn{1}{c}{\multirow{2}{*}{Accuracy(\%)}} \\
\cmidrule(lr){2-3} \cmidrule(lr){4-5}
\multicolumn{1}{l}{} & \multicolumn{1}{c}{$\phi_{w}^1$} & \multicolumn{1}{c}{$\phi_{w}^2$} & \multicolumn{1}{c}{$\phi_{s}^1$} & \multicolumn{1}{c}{$\phi_{s}^2$} & \multicolumn{1}{c}{} \\
\midrule
\multirow{4}{*}{Single Completer} & \multicolumn{1}{c}{\yes} &  &  &  & 78.2 \\
 &  & \multicolumn{1}{c}{\yes} &  &  & 80.1\\
 &  &  & \multicolumn{1}{c}{\yes} &  & 83.4 \\
 &  &  &  & \multicolumn{1}{c}{\yes} & 84.7 \\
\midrule
\multirow{6}{*}{Two Completers} & \multicolumn{1}{c}{\yes} & \multicolumn{1}{c}{\yes} &  &  & 82.3 \\
 &  &  & \multicolumn{1}{c}{\yes} & \multicolumn{1}{c}{\yes} & 85.6 \\
 & \multicolumn{1}{c}{\yes} &  & \multicolumn{1}{c}{\yes} &  & 94.1 \\
 & \multicolumn{1}{c}{\yes} &  &  & \multicolumn{1}{c}{\yes} & 93.8 \\
 &  & \multicolumn{1}{c}{\yes} & \multicolumn{1}{c}{\yes} &  & 95.2 \\
 &  & \multicolumn{1}{c}{\yes} &  & \multicolumn{1}{c}{\yes} & 94.7 \\
\bottomrule
\end{tabular}
\label{tab:exp_completer}
\end{table}
\subsection{Benchmarks}
\label{sec:appendix_benchmark_detail}
We provide more details about used benchmarks in Table~\ref{tab:benchmark_detail}. 
\begin{table}[htbp]
\centering
\caption{Details of all benchmarks used and corresponding metrics.}
    \begin{tabular}{lccc}
    \toprule
    \textbf{Benchmark} & \textbf{Split} & \textbf{\# Samples} & \textbf{Metric} \\
    \midrule
    MathVista~\citep{lu2024mathvista} & Testmini & 1000 & Accuracy \\
    MathVision~\citep{mathvision} & Full & 3040 & Accuracy \\
    MathVerse~\citep{zhang2024mathverse} & Vision-Only & 788 & Accuracy \\
    DynaMath~\citep{zou2025dynamath} & Full & 5050 & Accuracy \\
    WeMath~\citep{wemath} & Testmini & 1740 & Score (Strict)\\
    LogicVista~\citep{xiao2024logicvista} & Full & 448 & Accuracy\\
    MMMU~\citep{yue2024mmmu} & Validation & 900 & Accuracy \\  
    GSM8K~\citep{cobbe2021training} & Test & 1319 & Accuracy \\
    MATH~\citep{hendrycks_math} & Test & 5000 & Accuracy \\
    VisualProcessBench~\citep{wang2025visualprm} & - &  2866  & F1-score \\
    \bottomrule
    \end{tabular}
\label{tab:benchmark_detail}
\end{table}
\subsection{Reward Choice}
\label{sec:appendix_reward_choice}
In this paper, we consider three reward choices of \ours-PRM for aggregating step scores into a final score~\citep{zhang2025lessons}, including PRM-minimum, PRM-last, and PRM-product. For all reasoning steps $\{a_i\}_{i=1}^K$, its corresponding reward for each step is $\{r_i\}_{i=1}^K$, the reward $r$ for each solution with different scoring methods are calculated as follows: 1) PRM-minimum uses the minimum reward across all steps, i.e., $r=\min\{r_i\}_{i=1}^{K}$; 2) PRM-last uses the reward of last step, i.e., $r=r_K$; 3) PRM-product uses the reward products of all steps, i.e. $r=\prod_{i=1}^{K}r_i$. 

We list results in Table~\ref{tab:ablation_reward_choice}. It is shown that different choices of selecting a reward for PRMs could produce different performance. In all other experiments, we use the minimum reward across all steps as the reward for the solution unless otherwise specified.

\begin{table}[htbp]
\centering
\caption{Best-of-8 performance comparison on the MathVista~\citep{lu2024mathvista} with Qwen2.5-VL-7B~\citep{qwen2.5vl} with different reward choices: last step reward, minimum reward across all steps and product of all rewards. Default settings are marked in \colorbox{mygray}{gray}.}
    \begin{tabular}{lc}
    \toprule
    \textbf{Method} & \textbf{Accuracy(\%)} \\
    \midrule
    Zero-shot & 68.1 \\
    Self-consistency & 71.6 \\
    ORM & 72.8 \\
    PRM-product & 74.5 \\
    PRM-last & 75.1 \\
    \rowcolor{mygray} PRM-minimum & \textbf{75.2} \\
    \bottomrule
    \end{tabular}
\label{tab:ablation_reward_choice}
\end{table}

\subsection{Label Distribution}
We present the distribution of process labels in Table~\ref{tab:label_stats}. It is shown that \textit{label imbalance} exists in PRM800K and \ours-300K. We try different up-sample rates (e.g., $2\times$) for samples with error steps and find that slightly up-sampling negative data is beneficial for training PRMs, as shown in Table~\ref{tab:ablation_upsample}. We find that increasing the up-sample rate will not improve performance. We set the up-sample rate to 2 for other experiments.
\begin{table}[htbp]
\centering
\caption{Distribution of step labels across different datasets. Good, neutral, and bad denote different process labels.}
    \begin{tabular}{lccc}
    \toprule
    \textbf{Dataset} & \textbf{Good} & \textbf{Neutral} & \textbf{Bad} \\
    \midrule
    PRM800K~\citep{lightman2024lets} & 76\% & 12\% & 12\%   \\
    MC-300K & 81\% & - & 19\%  \\
    \ours-5K & 79\% & - & 21\% \\
    \bottomrule
    \end{tabular}
\label{tab:label_stats}
\end{table}
\begin{table}[htbp]
\centering
\caption{Results on MathVista under Best-of-8 setting with different up-sampling rate. We use \ours-5K as training dataset and Qwen-2.5-VL-7B as the policy model. Default settings are marked in \colorbox{mygray}{gray}.}
\begin{tabular}{lcc}
\toprule
\textbf{Method} & \textbf{Up-sample rate} &\textbf{Accuracy(\%)} \\
\midrule
Zero-shot & - & 68.1 \\
Self-consistency & -&  71.6 \\
\ours-ORM & - & 72.8 \\
\midrule
\ours-PRM & \no  & 74.8 \\
\rowcolor{mygray}\ours-PRM & $\times 2$   & \textbf{75.2} \\
\ours-PRM & $\times 5$   & \textbf{75.2} \\
\ours-PRM & $\times 10$   & 74.9 \\
\bottomrule
\end{tabular}
\label{tab:ablation_upsample}
\end{table}

\subsection{Effectiveness of Athena-PRM for Reward Ranked Fine-tuning}
Besides selecting responses with Athena-PRM, we have tried to directly use all responses to fine-tune. We do not use Athena-ORM to select responses because all data we use in reward-ranked fine-tuning has ground-truth answers. Before we use all data to fine-tune, we conduct a quick test to verify the effectiveness of response ranking with Athena-PRM. We used a small-scale dataset (30k queries) to verify the effectiveness of response ranking with Athena-PRM. Results are shown in Table~\ref{tab:ablation_raft}.

\begin{table}[htbp]
\centering
\caption{Results of reward ranked fine-tuning (RAFT) with different data settings. We report accuracy on the MathVista with Qwen2.5-VL-7B.}
    \begin{tabular}{lccc}
    \toprule
    \textbf{Method} & \textbf{Data} & \textbf{Athena-PRM} & \textbf{Accuracy} \\
    \midrule
    Qwen2.5-VL-7B & - & - & 68.1 \\
    +RAFT & 30k & \no & 68.7\\
    +RAFT & 30k & \yes & 69.8\\
    +RAFT (ours) &  150k & \yes & 71.4\\
    \bottomrule
    \end{tabular}
\label{tab:ablation_raft}
\end{table}

\subsection{Training Dynamics}
\begin{figure}
    \centering
    \includegraphics[width=0.7\textwidth]{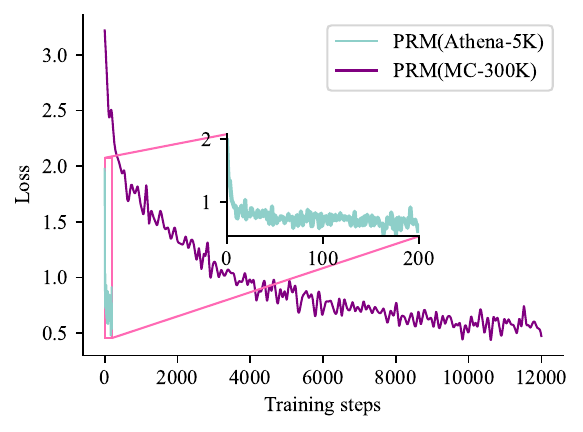}
    \caption{
    Training loss curve under different data.
    }
    \label{fig:loss}
\end{figure}

We show training loss curves for training PRMs with different datasets in Figure~\ref{fig:loss}. It is noticed that training PRMs with Athena-5K converges faster than MC-300K.

\subsection{Efficiency Discussion}
We measure efficiency using GPU hours for data synthesis and training PRMs. For data synthesis, i.e. estimate process labels, the proposed consistency strategy achieves 45$\times$ reduction in terms of GPU hours (10 hours on 8 AMD MI-250 GPUs) and we obtain 5K samples with process labels. For previous MC-based estimation methods, we labeled 300K samples which need 20 days on 8 AMD MI-250 GPUs. For training PRMs, using Athena-5k only requires 1/60 GPU hours compared with using MC-300K. As for the performance, the PRM training with Athena-5K get 75.2\% accuracy and the PRM trained with MC-300K gets 75.5\% accuracy under the best-of-8 setting on the MathVista, as shown in Table~\ref{tab:efficiency}.
\begin{table}[htbp]
\centering
\caption{GPU hours for data annotation and training.}
    \begin{tabular}{lcc}
    \toprule
     & Vanilla MC-based	& Consistency filter \\
    \midrule
    GPU hours (data annotation) & 20 days & 10 hours (\textbf{45$\times$} reduction) \\
    GPU hours (training) & 12 hours & 12 mins (\textbf{60$\times$} faster) \\
    MathVista accuracy (best-of-8) & 75.5 & 75.2 \\
    \bottomrule
    \end{tabular}
\label{tab:efficiency}
\end{table}

\subsection{PRMs in the Area of Thinking Models}
Recently, thinking models such as DeepSeek-R1~\citep{deepseekr1} have attracted a lot of attention in the research area, particularly for reinforcement learning with verifiable rewards. We conduct preliminary experiments to verify whether the proposed method could be adopted for thinking models, although we mainly conduct experiments with non-thinking models. We found that when we use thinking models such as \texttt{DeepSeek-R1-Distill-Qwen-7B} as the strong completer, most steps are estimated as correct steps. Thinking models usually have a stronger reflection ability than non-thinking models. Therefore, even if the intermediate steps are wrong, the final answers are correct in most cases. To be specific, more than 98\% of the steps are estimated as “correct” steps. We speculate that the pattern of thinking models and non-thinking models are very different. Further research is needed on whether verifying to explore whether to verify the correctness of intermediate steps for thinking models~\citep{multi-rm}.

\section{Case Study} 
\begin{figure}[htbp] 
    \centering
    \includegraphics[height=0.5\linewidth]{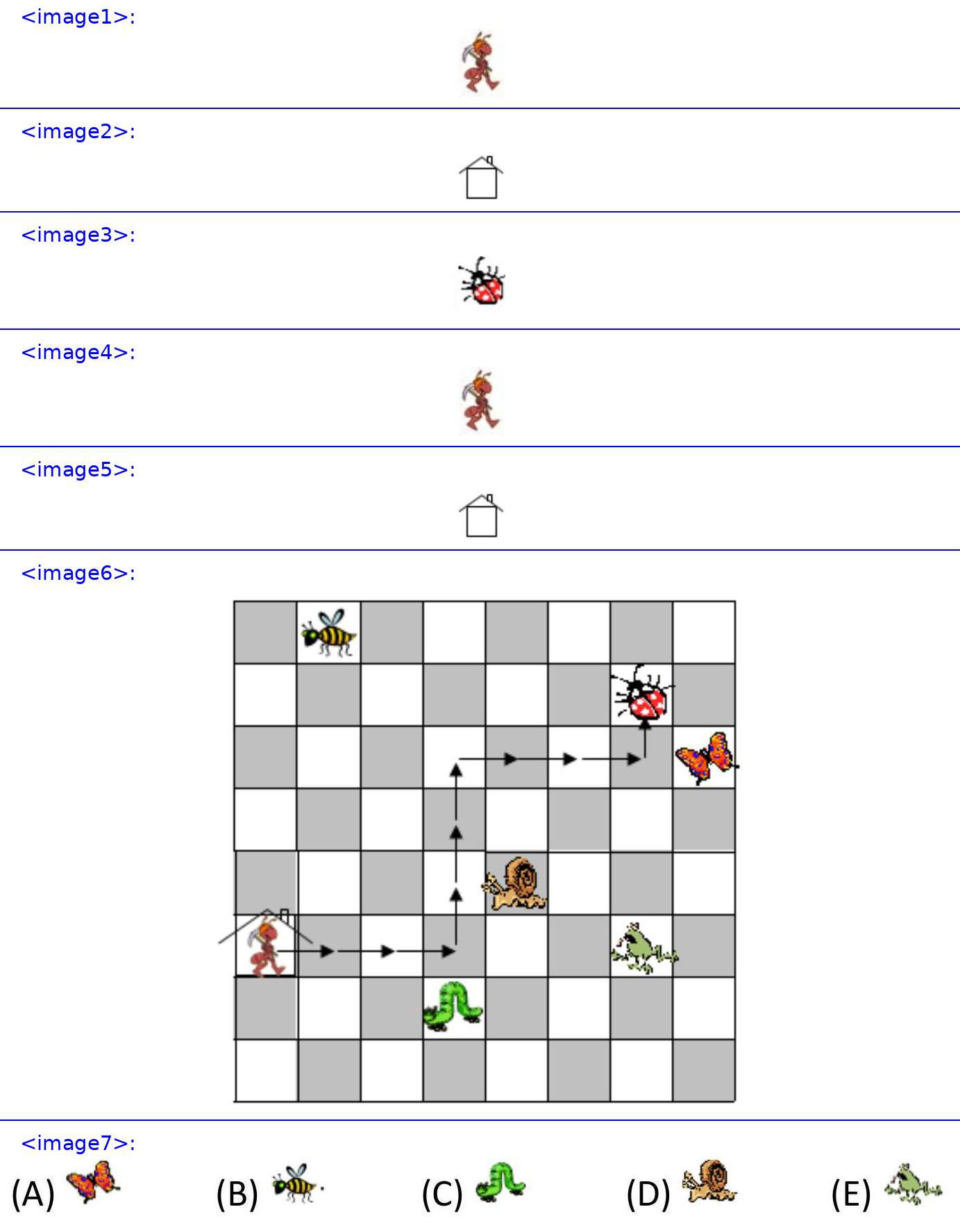}   
    \begin{tabular}{lcc}
    \toprule
    \multicolumn{3}{p{\linewidth}}{Hint: Please answer the question and provide the correct option letter, e.g., A, B, C, D, at the end.
Question: When the ant <image1> walks from home <image2> along the arrows $\rightarrow 3, \uparrow 3, \rightarrow 3, \uparrow 1$, he gets to the ladybird <image3>.

Which animal does the ant <image1> get to when he walks from home <image2> along the following arrows: $\rightarrow 2, \downarrow 2, \rightarrow 3, \uparrow 3, \rightarrow 2, \uparrow 2$?

<image6>

<image7>

Choices:(A) A (B) B (C) C (D) D (E) E

(Answer: A)
} \\
    \midrule
    Step & Reward & Label \\
    \midrule
    \parbox[t]{0.9\linewidth}{
    To determine which animal the ant reaches, we will follow the given movements step-by-step on the grid:}   & 0.95 & \textcolor{darkergreen}{\ding{52}} \\
    \midrule
    1. Start at the home position (`<image2>`), which is at the bottom-left corner of the grid.   & 0.99 & \textcolor{darkergreen}{\ding{52}} \\
    \midrule
    \parbox[t]{0.9\linewidth}{
        2. Follow the arrows:\\
        - $\rightarrow 2$: Move 2 squares to the right.\\
        - $\downarrow 2$: Move 2 squares down.\\
        - $\rightarrow 3$: Move 3 squares to the right.\\
        - $\uparrow 3$: Move 3 squares up.\\
        - $\rightarrow 2$: Move 2 squares to the right.\\
        - $\uparrow 2$: Move 2 squares up.
    }   & 0.99 & \textcolor{darkergreen}{\ding{52}}\\  
    \midrule
    \parbox[t]{0.9\linewidth}{
        3. Trace the movement:\\
        - Start at the bottom-left corner.\\
        - Move 2 squares right: You are now in the third column of the first row.\\
        - Move 2 squares down: You are now in the third column of the third row.\\
        - Move 3 squares right: You are now in the sixth column of the third row.\\
        - Move 3 squares up: You are now in the sixth column of the top row.\\
        - Move 2 squares right: You are now in the eighth column of the top row.\\
        - Move 2 squares up: You are now in the eighth column of the top row (still at the butterfly position). 
    } & 0.12 & \textcolor{red2}{\ding{56}} \\
    \midrule
    \parbox[t]{0.9\linewidth}{
    The ant reaches the \textbf{butterfly}, which corresponds to \textbf{option A}.} & 0.87 & \textcolor{darkergreen}{\ding{52}} \\
    \midrule
    \parbox[t]{0.9\linewidth}{
    \#\#\# Final Answer: \textbf{A} } & 0.75 & \textcolor{darkergreen}{\ding{52}} \\
    \bottomrule
\end{tabular}
\vspace{-.5em}
    \caption{A case study from VisualProcessBench~\citep{wang2025visualprm}.}
    \label{fig:case_1}
\end{figure}

\begin{figure}[t]
    \centering
    \includegraphics[width=0.3\linewidth]{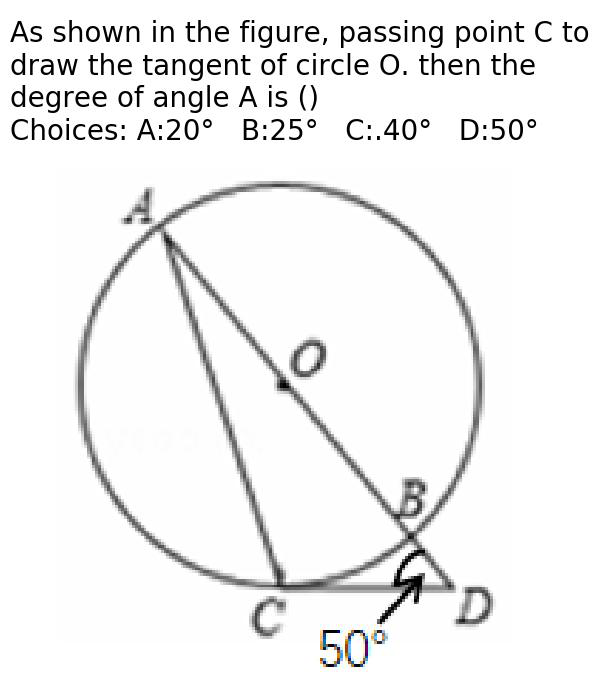}
    
    \begin{tabular}{lcc}
    \toprule
    \multicolumn{3}{p{\linewidth}}{According to the question shown in the image, please directly answer the question and provide the correct option letter, e.g., A, B, C, D.

(Answer: A)
} \\
    \midrule
    Step & Reward & Label \\
    \midrule
    \parbox[t]{0.9\linewidth}{
    To find the degree of angle A, we can use the property that the angle subtended by an arc at the center of the circle is twice the angle subtended at any point on the circle.}   & 0.82 &  \textcolor{darkergreen}{\ding{52}} \\
    \midrule
    \parbox[t]{0.9\linewidth}{
    1. \textbf{Identify the key points:}\\   
    - The arcs are AC and BC.\\   
    - Angle A is subtended by these arcs at point O (the center of the circle). }  & 0.85 &  \textcolor{darkergreen}{\ding{52}} \\
    \midrule
    \parbox[t]{0.9\linewidth}{
       2. \textbf{Understand the given information:}\\
       - CD is a tangent at point C.\\  
       - The angle between the tangent and the chord is equal to the angle subtended by the chord at the circumference. Therefore, angle CDB = 50°.
    }   & 0.85 &   \textcolor{darkergreen}{\ding{52}}  \\  
    \midrule
    \parbox[t]{0.9\linewidth}{
        3. \textbf{Angle properties:}\\
        - The angle A is subtended by arc BC at the center.\\
        - Since CD is a tangent, angle A = 50° (as the inscribed angle is half the central angle). 
    } & 0.23 & \textcolor{red2}{\ding{56}} \\
    \midrule
    \parbox[t]{0.9\linewidth}{
    Thus, the degree of angle A is 20°.} & 0.67 &  \textcolor{darkergreen}{\ding{52}}  \\
    \midrule
    \parbox[t]{0.9\linewidth}{
    Final answer: A} & 0.96 & \textcolor{darkergreen}{\ding{52}} \\
    \bottomrule
\end{tabular}    
    \caption{A case study from VisualProcessBench~\citep{wang2025visualprm}.}
    \label{fig:case_2}
\end{figure}

\begin{figure}[htbp]
    \centering
    \includegraphics[height=0.4\linewidth]{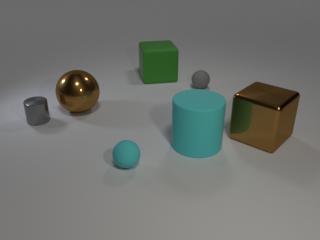}
    
    \begin{tabular}{lcc}
    \toprule
    \multicolumn{3}{p{\linewidth}}{Hint: Please answer the question requiring an integer answer and provide the final value, e.g., 1, 2, 3, at the end. 
    
Question: Subtract all large green matte cubes. Subtract all big green blocks. How many objects are left?

(Answer: 6)
} \\
    \midrule
    Step & Reward & Label \\
    \midrule
    \parbox[t]{0.9\linewidth}{
    To solve this problem, let's identify and count the objects in the image:}   &  0.95
 &  \textcolor{darkergreen}{\ding{52}} \\
    \midrule
    \parbox[t]{0.9\linewidth}{
   1. There is one large green matte cube.}  & 0.99 &  \textcolor{darkergreen}{\ding{52}} \\
    \midrule
    \parbox[t]{0.9\linewidth}{
     2. There are no other big green blocks since the only green block is already counted.}   & 0.90 &   \textcolor{darkergreen}{\ding{52}}  \\  
    \midrule    
    \parbox[t]{0.9\linewidth}{
        Now, we subtract these objects as per the instructions} & 0.92 &\textcolor{darkergreen}{\ding{52}} \\
    \midrule
    
    \parbox[t]{0.9\linewidth}{
  - Subtract all large green matte cubes: This removes the large green matte cube from the count.} & 0.99 & \textcolor{darkergreen}{\ding{52}} \\
    \midrule
    
    \parbox[t]{0.9\linewidth}{
    - Subtract all big green blocks: Since there are no other big green blocks, nothing additional is removed.} & 0.99 & \textcolor{darkergreen}{\ding{52}} \\
    \midrule
    \parbox[t]{0.9\linewidth}{
    After removing the large green matte cube, the remaining objects are:\\
    - A small gray metallic cylinder.\\
    - A large gold metallic sphere.\\
    - A small gray matte sphere.\\
    - A large cyan matte cylinder.\\
    - A small cyan matte sphere.\\
    - A large gold metallic cube. } & 0.84 & \textcolor{darkergreen}{\ding{52}} \\

    \midrule
    \parbox[t]{0.9\linewidth}{
    Counting the remaining objects, we have 6 objects left.} & 0.92 & \textcolor{darkergreen}{\ding{52}} \\

    \midrule
    \parbox[t]{0.9\linewidth}{
    Answer: $\boxed{6}$ } & 0.99 & \textcolor{darkergreen}{\ding{52}} \\
    \bottomrule
\end{tabular}    
    \caption{A case study from MathVista~\citep{lu2024mathvista} with Qwen2.5-VL-72B~\citep{qwen2.5vl}.}
    \label{fig:case_5}
\end{figure}

We provide some examples from VisualProcessBench and MathVista to help readers understand how PRMs work, see Figure~\ref{fig:case_1}, ~\ref{fig:case_2}, ~\ref{fig:case_4}, ~\ref{fig:case_5}.
\begin{figure}[htbp]
    \centering
    \includegraphics[height=0.4\linewidth]{fig/example_fig/mathvista_559.jpg}
    
    \begin{tabular}{lcc}
    \toprule
    \multicolumn{3}{p{\linewidth}}{Hint: Please answer the question requiring an integer answer and provide the final value, e.g., 1, 2, 3, at the end. 
    
Question: what is the highest number shown on the black outer part of the watch?

(Answer: 55)
} \\
    \midrule
    Step & Reward & Label \\
    \midrule
    \parbox[t]{0.9\linewidth}{
    To determine the highest number shown on the black outer part of the watch, let's analyze the image step by step}   &  0.90
 &  \textcolor{darkergreen}{\ding{52}} \\
    \midrule
    \parbox[t]{0.9\linewidth}{
   1. The black outer part of the watch appears to be a bezel with numbers printed on it. }  & 0.87 &  \textcolor{darkergreen}{\ding{52}} \\
    \midrule
    \parbox[t]{0.9\linewidth}{
      2. The numbers on the bezel are arranged in a circular pattern.}   & 0.96 &   \textcolor{darkergreen}{\ding{52}}  \\  
    \midrule    
    \parbox[t]{0.9\linewidth}{
        3. Observing the numbers, they range from 0 to 60, increasing in increments of 5.} & 0.83 &\textcolor{darkergreen}{\ding{52}} \\
    \midrule
    
    \parbox[t]{0.9\linewidth}{
   4. The highest number visible on the bezel is 60.} & 0.34 &  \textcolor{red2}{\ding{56}} \\
    \midrule
    
    \parbox[t]{0.9\linewidth}{
    Therefore, the highest number shown on the black outer part of the watch is 60.} & 0.15 & \textcolor{red2}{\ding{56}} \\
    \midrule
    \parbox[t]{0.9\linewidth}{
    Answer: $\boxed{60}$ } & 0.04 & \textcolor{red2}{\ding{56}} \\
    
    \bottomrule
\end{tabular}    
    \caption{A case study from MathVista~\citep{lu2024mathvista} with Qwen2.5-VL-72B~\citep{qwen2.5vl}.}
    \label{fig:case_4}
\end{figure}

\end{document}